\documentclass{article}

\usepackage[preprint]{colm2026_conference}

\usepackage{microtype}
\usepackage{needspace}
\usepackage{hyperref}
\usepackage{url}
\usepackage{booktabs}
\usepackage{lineno}
\usepackage{float}
\usepackage{enumitem}
\usepackage{longtable}
\usepackage{graphicx}
\usepackage{subcaption}
\usepackage{pgf-pie}
\usepackage{amsmath}
\usepackage{amssymb}
\usepackage{mathtools}
\usepackage{amsthm}
\usepackage[capitalize,noabbrev]{cleveref}
\usepackage[textsize=tiny]{todonotes}
\usepackage{tikz}
\usetikzlibrary{decorations.pathreplacing}
\usetikzlibrary{backgrounds}
\usetikzlibrary{arrows.meta}
\usetikzlibrary{positioning}
\usetikzlibrary{fadings}
\usetikzlibrary{positioning, shapes.geometric, arrows.meta, fit}
\usepackage[most]{tcolorbox}
\usetikzlibrary{calc}
\definecolor{darkblue}{rgb}{0, 0, 0.5}
\hypersetup{
    colorlinks=true,
    citecolor=darkblue,
    linkcolor=darkblue,
    urlcolor=darkblue
}

\theoremstyle{plain}

\theoremstyle{definition}

\theoremstyle{remark}

\newcommand{\ourbenchmark}{{\sf StyleBench}}

\definecolor{leftcolor}{RGB}{251,188,174}
\definecolor{leftbg}{RGB}{252,248,240}
\definecolor{lefttitle}{RGB}{84,45,45}

\definecolor{middlecolor}{RGB}{178,235,242}
\definecolor{middlebg}{RGB}{240,253,248}
\definecolor{middletitle}{RGB}{26,82,82}

\definecolor{rightcolor}{RGB}{221,214,254}
\definecolor{rightbg}{RGB}{250,245,255}
\definecolor{righttitle}{RGB}{65,44,86}

\newtcolorbox{conclusionbox}{
    colback=blue!5!white,
    colframe=blue!75!black,
    boxrule=0.8pt,
    arc=2mm,
    left=3mm, right=3mm, top=2mm, bottom=2mm
}

\newtcolorbox{llmresponse}[1][]{
    enhanced,
    breakable,
    colback=blue!3,
    colframe=blue!55!black,
    boxrule=0.5pt,
    arc=2pt,
    left=5pt,right=5pt,top=4pt,bottom=4pt,
    title={LLM Response},
    fonttitle=\bfseries,
    before skip=4pt,
    after skip=4pt,
    boxsep=1pt,
    parbox=false,
    #1
}

\newtcolorbox{game24response}[1][]{
    enhanced,
    breakable,
    colback=green!3,
    colframe=green!50!black,
    boxrule=0.5pt,
    arc=2pt,
    left=5pt,right=5pt,top=4pt,bottom=4pt,
    title={LLM Response: Game24},
    fonttitle=\bfseries,
    before skip=4pt,
    after skip=4pt,
    boxsep=1pt,
    parbox=false,
    #1
}

\newtcolorbox{comprehensivegame24}[1][]{
    enhanced,
    breakable,
    colback=green!3,
    colframe=green!50!black,
    boxrule=0.5pt,
    arc=2pt,
    left=5pt,right=5pt,top=4pt,bottom=4pt,
    title={LLM Responses: Game24 -- Multiple Reasoning Styles},
    fonttitle=\bfseries,
    before skip=4pt,
    after skip=4pt,
    boxsep=1pt,
    parbox=false,
    #1
}

\newtcolorbox{commonsenseqa}[1][]{
    enhanced,
    breakable,
    colback=purple!3,
    colframe=purple!50!black,
    boxrule=0.5pt,
    arc=2pt,
    left=5pt,right=5pt,top=4pt,bottom=4pt,
    title={LLM Responses: CommonsenseQA -- Reasoning Style Comparison},
    fonttitle=\bfseries,
    before skip=4pt,
    after skip=4pt,
    boxsep=1pt,
    parbox=false,
    #1
}
\setlist[itemize]{topsep=2pt,itemsep=1pt,parsep=0pt,partopsep=0pt,leftmargin=1.2em}
\setlist[enumerate]{topsep=2pt,itemsep=1pt,parsep=0pt,partopsep=0pt,leftmargin=1.4em}

\renewenvironment{quote}{%
  \list{}{%
    \leftmargin=2em\rightmargin=2em%
    \topsep=4pt\parsep=0pt\partopsep=0pt%
  }\item\relax
}{\endlist}

\title{StyleBench: Evaluating  thinking styles in Large Language Models}

\author{
  Junyu Guo\thanks{Corresponding author: \texttt{junyuguo24@berkeley.edu}} \\
  University of California, Berkeley \\
  \texttt{}
  \And
  Shangding Gu \\
  University of California, Berkeley \\
  \texttt{}
  \And
  Ming Jin \\
  Virginia Tech \\
  \texttt{}
  \And
  Costas Spanos \\
  University of California, Berkeley \\
  \texttt{}
  \And
  Javad Lavaei \\
  University of California, Berkeley \\
  \texttt{}
}

\begin{document}

\ifcolmsubmission
\linenumbers
\fi

\maketitle

\begin{abstract}
Structured reasoning can improve the inference performance of large language models (LLMs), but it also introduces computational cost and control constraints. When additional reasoning structure helps, and when it instead reduces efficiency or robustness, remains poorly understood. We propose \ourbenchmark, where we  study reasoning structure as a capacity-constrained design choice rather than a fixed inference recipe. We evaluate five representative reasoning styles---Chain-of-Thought, Tree-of-Thought, Algorithm-of-Thought, Sketch-of-Thought, and Chain-of-Draft---across five reasoning tasks and 15 open-source LLMs ranging from 270M to 120B parameters. We find that greater structural complexity improves accuracy only in limited regimes defined by task demands and model capacity. Search-based styles help on open-ended combinatorial problems but fail on smaller models, while concise styles achieve large efficiency gains on structured tasks without sacrificing performance. We also identify systematic failure modes in smaller models, including premature guessing and weak adherence to reasoning-control instructions. To study adaptive reasoning control, we further compare supervised and reinforcement-based strategy selection on Qwen-7B-Instruct. Supervised fine-tuning collapses to shallow style preferences, whereas GRPO learns stronger adaptive control and improves downstream performance. Together, these results clarify when structured reasoning is useful, when it is wasteful, and why learning to choose a reasoning strategy is itself a challenging inference problem,  we open source the benchmark in \url{https://github.com/JamesJunyuGuo/Style_Bench}.
\end{abstract}

\section{Introduction}
Large Language Models (LLMs) have demonstrated impressive capabilities across a diverse range of tasks, including mathematical reasoning, code generation, and complex question answering~\citep{imani2023mathprompter, wang2023review, tan2023can}. A key insight from prior work is that their performance on challenging problems is not merely a function of scale, but is critically dependent on the methods used to guide reasoning~\citep{huang2025gemini}. This has led to a range of prompting techniques and test-time methods that structure a model’s reasoning process, providing a diverse set of tools for probing how different forms of reasoning affect inference performance.
 Notable among these are \textit{Chain-of-Thought} (CoT)~\citep{wei2022chain}, which decomposes problems into sequential steps, and more advanced paradigms like \textit{Tree-of-Thought} (ToT)~\citep{yao2023tree}, which explores multiple reasoning paths in parallel, and \textit{Reasonflux}~\citep{yang2025reasonflux}, employing high-level templates to explore potential solutions.

Despite the impressive capabilities of recent large-scale models such as GPT-4o~\citep{agarwal2025gpt}, DeepSeek~\citep{guo2025deepseek}, Gemini-2.5~\citep{comanici2025gemini}, and Qwen3~\citep{yang2025qwen3}, their performance on specialized or highly complex tasks continues to exhibit critical limitations. Performance is often sensitive to how prompts are phrased, and models usually need iterative feedback to get reliable results~\citep{sel2023algorithm,yao2023tree,renze2024self}. In response, recent work has sought to automate reasoning strategy selection. For instance, \cite{gao2024meta,guo2025meta} proposed a two-stage meta-reasoning process to identify the optimal style, while \cite{wan2025rema} leveraged generated instructions to guide the model. However, these approaches assume the existence of a well-understood mapping between problem types and optimal reasoning strategies—a foundation we currently lack. The existing literature provides an incomplete picture, as evaluations are typically limited to a single reasoning style, a narrow set of tasks, or a small selection of models. This leaves a significant gap in our understanding of how these strategies generalize across different model architectures, problem domains, and computational budgets. Importantly, the goal of this work is not to advocate for the superiority of any particular reasoning strategy. Instead, we treat existing prompting and test-time reasoning methods as analytical instruments for understanding why certain forms of structured reasoning succeed in some regimes and fail in others. By examining how different reasoning structures interact with task complexity and model capacity, we aim to uncover the underlying factors that govern reasoning effectiveness, rather than  to identify the best-performing strategy. A central challenge in structured reasoning is the trade-off between reasoning depth and efficiency. For simple queries, effective inference often requires only a concise and direct response. Enforcing verbose reasoning chains, such as step-by-step decomposition in Chain of Thought (CoT)~\citep{wei2022chain}, can introduce unnecessary computational overhead in these cases. In contrast, complex problems frequently require more elaborate reasoning procedures to achieve high accuracy. These procedures may involve symbolic abstraction~\citep{aytes2025sketch} or iterative reasoning that reflects on intermediate results~\citep{renze2024self}. This tension raises a fundamental question. How can language models avoid overthinking on simple tasks while still reasoning effectively on more challenging ones~\citep{chen2024not,fang2025thinkless,sui2025stop}? An ideal reasoning strategy should adapt its cognitive load to the complexity of the problem rather than applying the same level of reasoning indiscriminately.




These gaps in our understanding of generalization and efficiency motivate the central question of this work. Rather than asking which reasoning strategy is universally superior, we seek to understand how and why different forms of structured reasoning succeed or fail across tasks, model scales, and architectures. In particular, we focus on the factors that govern the trade-off between reasoning accuracy and computational efficiency.

\begin{quote}
\textit{What aspects of reasoning structure contribute to successful inference in large language models, and what aspects instead introduce inefficiency or systematic failure under different task and capacity regimes?}
\end{quote}

To address this question, we propose \ourbenchmark, where we conduct a systematic empirical analysis using a diverse set of existing reasoning strategies as analytical probes. By examining their behavior across multiple tasks and model scales, we uncover consistent patterns, capacity dependent effects, and failure modes in LLM reasoning. Our goal is not to rank or recommend specific strategies, but to reveal the underlying phenomena that shape when structured reasoning helps, when it is unnecessary, and when it breaks down, thereby providing insights that extend beyond any single method.

Our contributions can be summarized as follows:
\begin{itemize}
    \item \textbf{A systematic view of reasoning structure.} We conceptualize reasoning strategies as capacity-constrained design choices, and  demonstrate that increasing reasoning structure introduces a trade-off between accuracy and efficiency.
    
    \item \textbf{Empirical evidence of a capacity-dependent reasoning trade-off.} Through a systematic study across diverse tasks and model scales, we identify consistent regimes in which structured reasoning improves performance, as well as regimes where it degrades robustness or incurs unnecessary computational cost.
    
    \item \textbf{Failure modes of over-structured reasoning.} We characterize how search-based and heavily structured reasoning strategies fail on smaller or misaligned models, exhibiting error amplification and  instruction-following breakdowns.
    
    \item \textbf{Limits and possibilities of learned reasoning control.} We show that SFT for reasoning structure selection collapses to superficial style preferences, pointing to a gap between generating reasoning and controlling it. In contrast, GRPO yields improved adaptive control on Qwen-7B-Instruct, suggesting that online credit assignment is a more effective path toward learning when and how to reason.
\end{itemize}

\begin{figure}[htbp]
\centering
\small

\begin{minipage}[t]{0.485\linewidth}
\begin{tcolorbox}[
    enhanced,
    colback=blue!4,
    colframe=blue!55!black,
    colbacktitle=blue!55!black,
    coltitle=white,
    boxrule=0.6pt,
    arc=2.5pt,
    left=4pt,right=4pt,top=4pt,bottom=4pt,
    title=\textbf{Example 1: Step-by-Step Reasoning (CoT)},
    fonttitle=\bfseries\small,
    height=2.8cm
]
\textbf{Problem:} Solve $2x+5=15$.

\vspace{2pt}
\textbf{Style:} CoT works through each step explicitly in sequence.

\vspace{2pt}
\textbf{Response:} Step 1: subtract 5 $\Rightarrow$ $2x=10$. Step 2: divide by 2 $\Rightarrow$ $x=5$.
\end{tcolorbox}
\end{minipage}\hfill
\begin{minipage}[t]{0.485\linewidth}
\begin{tcolorbox}[
    enhanced,
    colback=teal!4,
    colframe=teal!55!black,
    colbacktitle=teal!55!black,
    coltitle=white,
    boxrule=0.6pt,
    arc=2.5pt,
    left=4pt,right=4pt,top=4pt,bottom=4pt,
    title=\textbf{Example 2: Symbolic Sketch (SoT)},
    fonttitle=\bfseries\small,
    height=2.8cm
]
\textbf{Problem:} All birds have wings. Penguins are birds. Wings?

\vspace{2pt}
\textbf{Style:} SoT encodes the argument as a compact symbolic sketch, skipping verbose prose.

\vspace{2pt}
\textbf{Response:} $B\!\subseteq\!W,\;P\!\subseteq\!B \;\Rightarrow\; P\!\subseteq\!W$. Yes.~$\checkmark$
\end{tcolorbox}
\end{minipage}

\vspace{4pt}

\begin{minipage}[t]{0.485\linewidth}
\begin{tcolorbox}[
    enhanced,
    colback=violet!4,
    colframe=violet!55!black,
    colbacktitle=violet!55!black,
    coltitle=white,
    boxrule=0.6pt,
    arc=2.5pt,
    left=4pt,right=4pt,top=4pt,bottom=4pt,
    title=\textbf{Example 3: Branch \& Vote (ToT)},
    fonttitle=\bfseries\small,
    height=3.8cm
]
\textbf{Problem:} Bat+ball=\$1.10. Bat costs \$1 more. Ball=?

\vspace{2pt}
\textbf{Style:} ToT explores branches; majority vote picks the answer.

\vspace{2pt}
\textbf{Response:}
Branch~1: \$0.10 $\to$ total \$1.20~$\times$\quad
Branch~2: \$0.05 $\to$ total \$1.10~$\checkmark$\quad
Branch~3: \$0.05~$\checkmark$

Vote (2/3): \textbf{\$0.05}
\end{tcolorbox}
\end{minipage}\hfill
\begin{minipage}[t]{0.485\linewidth}
\begin{tcolorbox}[
    enhanced,
    colback=orange!5,
    colframe=orange!65!black,
    colbacktitle=orange!65!black,
    coltitle=white,
    boxrule=0.6pt,
    arc=2.5pt,
    left=4pt,right=4pt,top=4pt,bottom=4pt,
    title=\textbf{Example 4: Try \& Backtrack (AoT)},
    fonttitle=\bfseries\small,
    height=3.8cm
]
\textbf{Problem:} Find path S$\!\to\!$G: edges S$\!\to\!$A, A$\!\to\!$B (dead end), A$\!\to\!$C, C$\!\to\!$G.

\vspace{2pt}
\textbf{Style:} AoT tries paths and \textbf{backtracks} when stuck.

\vspace{2pt}
\textbf{Response:}
Try S$\!\to\!$A$\!\to\!$B: dead end $\to$ \textbf{backtrack} to A.

Try S$\!\to\!$A$\!\to\!$C$\!\to\!$G: success.~$\checkmark$
\end{tcolorbox}
\end{minipage}

\caption{Four reasoning styles, each illustrated by its defining characteristic: CoT chains explicit steps, SoT compresses reasoning into symbolic sketches, ToT explores branches and selects by majority vote, and AoT backtracks when a path fails.}
\label{fig:reasoning_style_examples}
\end{figure}

\section{Preliminary}\label{sec:preliminary}

In this work, we study \emph{reasoning structure} as a property of inference rather than as a collection of isolated prompting techniques. While prior work often presents reasoning strategies as standalone methods, our goal is to understand the structural factors that govern why such strategies succeed in some settings and fail in others. To this end, we treat existing reasoning strategies as concrete instantiations of broader design choices that shape how intermediate reasoning is produced and represented during inference. Formally, we consider an inference process in which a large language model $f_\theta$ maps an input $x \in \mathcal{X}$ to an output $y \in \mathcal{Y}$ through a sequence of intermediate reasoning states. A reasoning strategy specifies how this sequence is generated and constrained during inference. Our analysis focuses on how different structural choices in this process affect both performance and efficiency.

Importantly, our analysis does not aim to compare or rank reasoning strategies as end-to-end systems. Instead, we use them as controlled probes to expose how different forms of reasoning structure interact with task demands and model capacity. This perspective allows us to move beyond method level comparison and toward a systematic understanding of reasoning behavior in large language models.

\subsection{Structural Dimensions of Reasoning}

We model structured reasoning as a sequence of intermediate states
\begin{equation}
    x \rightarrow z_1 \rightarrow z_2 \rightarrow \cdots \rightarrow z_T \rightarrow y,\label{intermediate states}
\end{equation}
where each $z_t$ represents an intermediate reasoning representation generated during inference. Reasoning strategies differ in how these intermediate states are constructed, constrained, and consumed, leading to systematic differences in inference behavior and computational cost. We summarize several structural dimensions that are particularly relevant for understanding their success and failure modes.

\begin{figure}[t]
\centering
\begin{tikzpicture}[
    x=1cm,y=1cm,
    >=Stealth,
    line cap=round,
    line join=round,
    font=\small,
    main/.style={
        rectangle,
        rounded corners=3pt,
        draw=blue!65,
        fill=blue!10,
        thick,
        minimum width=1.15cm,
        minimum height=0.78cm,
        inner sep=2pt,
        align=center
    },
    branch/.style={
        rectangle,
        rounded corners=3pt,
        draw=orange!75!black,
        fill=orange!10,
        thick,
        minimum width=1.15cm,
        minimum height=0.78cm,
        inner sep=2pt,
        align=center
    },
    io/.style={
        rectangle,
        rounded corners=3pt,
        draw=green!50!black,
        fill=green!15,
        thick,
        minimum width=1.0cm,
        minimum height=0.78cm,
        inner sep=2pt,
        font=\small\bfseries,
        align=center
    },
    mainarrow/.style={->, thick, blue!70},
    brancharrow/.style={->, thick, dashed, orange!80!black},
    ann/.style={font=\scriptsize, align=center, text=black}
]

\node[io]    (x)  at (0,0) {$x$};
\node[main]  (z1) at (1.5,0) {$z_1$};
\node[main]  (z2) at (3.0,0) {$z_2$};
\node         (d)  at (4.25,0) {$\cdots$};
\node[main]  (zT) at (5.5,0) {$z_T$};
\node[io]    (y)  at (7.0,0) {$y$};

\draw[mainarrow] (x) -- (z1);
\draw[mainarrow] (z1) -- (z2);
\draw[mainarrow] (z2) -- (d);
\draw[mainarrow] (d) -- (zT);
\draw[mainarrow] (zT) -- (y);

\node[branch] (z2p) at (3.0,-1.35) {$z_2'$};
\node[branch] (z3p) at (4.5,-1.35) {$z_3'$};
\node[branch] (z2pp) at (3.0,-2.45) {$z_2''$};

\draw[brancharrow] (z1.south) |- (z2p.west);
\draw[brancharrow] (z2p.east) -- (z3p.west);
\draw[brancharrow] (z3p.north) |- (zT.south);

\draw[brancharrow] (z1.south) |- (z2pp.west);
\draw[brancharrow] (z2pp.east) -| ([yshift=-2pt]zT.south);


\draw[decorate,decoration={brace,amplitude=4pt},thick,gray!70]
    ($(z1.north west)+(0,0.22)$) -- ($(zT.north east)+(0,0.22)$)
    node[midway, above=6pt, ann] {Reasoning depth $T$};

\draw[decorate,decoration={brace,mirror,amplitude=4pt},thick,gray!70]
    ([xshift=-0.35cm]z2.north west) -- ([xshift=-0.35cm]z2pp.south west)
    node[midway, left=7pt, ann] {Branching\\complexity};

\node[ann] at (7.0,-1.15) {Inference\\cost $\mathcal{C}$};

\node[ann, anchor=north] at (3.5,-3.35) {
    \textcolor{blue!70}{\rule{0.9em}{0.9em}}~Main path
    \hspace{1.2em}
    \textcolor{orange!80!black}{\rule{0.9em}{0.9em}}~Alternative branches
    \hspace{1.2em}
    \textcolor{green!50!black}{\rule{0.9em}{0.9em}}~Input/Output
};

\end{tikzpicture}
\caption{Reasoning during inference varies along  depth, branching, and cost.}
\label{fig:reasoning_structure}
\end{figure}

First, \emph{reasoning depth} refers to the length $T$ of the intermediate sequence. Shallow reasoning corresponds to small $T$, potentially with a direct mapping from input to output, while deeper reasoning introduces multi-step decomposition and intermediate conclusions. Second, \emph{branching complexity} characterizes whether inference follows a single trajectory $\{z_t\}_{t=1}^T$ or explores a set of alternative candidates at each step. Branching or search-based strategies can increase coverage for open ended problems, but also amplify inference cost and error propagation. Third, \emph{intermediate representation density} refers to how much information is packed into each $z_t$. Some strategies expose full natural language explanations, while others rely on partial or compressed representations. This choice affects both interpretability and the accumulation of noise across reasoning steps. Fourth, \emph{control complexity} captures how strongly the reasoning process constrains model behavior. Highly structured prompts restrict the space of admissible intermediate states, which can guide inference but also increase the likelihood of instruction following failures under limited model capacity. Finally, \emph{inference cost} measures the computational resources consumed by the reasoning process, such as total token usage or latency. Generally, deeper reasoning, higher branching, and stricter control constraints induce higher inference cost. These dimensions do not cover every possible variation of reasoning structure. Rather, they provide a useful abstraction for analyzing how different reasoning strategies instantiate common structural design choices.

\subsection{A Trade-Off Between Accuracy and Efficiency}

Introducing additional reasoning structure can improve accuracy by decomposing complex problems into manageable subproblems. However, this benefit is not unconditional. Increased structure also leads to longer reasoning trajectories, higher inference cost, and tighter control constraints, which can degrade performance when model capacity or task complexity is misaligned. From this perspective, we view reasoning structure as a capacity-constrained design choice that induces a trade-off between accuracy and efficiency. Let $\mathcal{C}(z_{1:T})$ denote the inference cost associated with a reasoning sequence and $\mathcal{A}(y, y^\ast)$ denote task performance with respect to a ground truth $y^\ast$. Increasing reasoning depth or control may improve $\mathcal{A}$ in some regimes, but can also increase $\mathcal{C}$ without corresponding gains, or even reduce $\mathcal{A}$ due to error accumulation.

This observation leads to a central hypothesis that guides our empirical study. Increasing reasoning structure improves inference only within specific regimes defined by task demands and model capacity. Outside these regimes, additional reasoning may be unnecessary or harmful. In the following sections, we use empirical evidence to characterize when structured reasoning helps, when it is redundant or introduces systematic failure modes.

\section{Methodology}

Our empirical study examines how different forms of reasoning structure shape inference behavior in large language models. Rather than viewing reasoning strategies as competing prompt templates, we treat them as controlled variations of how intermediate reasoning is organized during inference. This perspective allows us to isolate how reasoning structure interacts with task complexity, model capacity, and computational cost under consistent evaluation settings. In addition to comparing fixed reasoning styles, we further study whether reasoning strategy selection itself can be learned, using both supervised fine-tuning (SFT) and reinforcement learning via GRPO.

Formally, let $x \in \mathcal{X}$ denote an input problem, $y \in \mathcal{Y}$ the final answer, and $z_{1:T}$ a sequence of intermediate reasoning states. A reasoning strategy determines how the model maps
\begin{equation}
x \rightarrow z_{1:T} \rightarrow y.
\end{equation}
Different strategies impose different structural constraints on the form, depth, and branching behavior of $z_{1:T}$, thereby affecting both task performance and inference cost.

\subsection{Reasoning Strategies}

We consider five representative reasoning strategies that span a range of structural properties, including reasoning depth, branching behavior, intermediate representation density, and control complexity. These strategies are selected to cover different regions of the reasoning design space rather than to cover every possible technique.

\textbf{Chain-of-Thought (CoT)}~\citep{wei2022chain} encourages models to decompose problems into a linear sequence of intermediate steps, producing a full natural language reasoning trace.
\textbf{Chain-of-Draft (CoD)}~\citep{xu2025chain} constrains this process by requiring concise and symbolic intermediate representations, reducing verbosity while retaining stepwise structure.
\textbf{Sketch-of-Thought (SoT)}~\citep{aytes2025sketch} further compresses intermediate reasoning by encouraging lightweight, task-specific symbolic representations.
In contrast, \textbf{Tree-of-Thought (ToT)}~\citep{yao2023tree} and \textbf{Algorithm-of-Thought (AoT)}~\citep{sel2023algorithm} introduce branching and backtracking mechanisms, enabling multiple reasoning trajectories to be explored during inference.

Let $s \in \mathcal{S}$ denote a reasoning style, where
\begin{equation}
\mathcal{S} = \{\text{CoT}, \text{CoD}, \text{SoT}, \text{ToT}, \text{AoT}\}.
\end{equation}
Under a fixed strategy $s$, inference follows
\begin{equation}
y \sim p_{\theta}(y \mid x, s),
\end{equation}
where the prompt template associated with $s$ determines how intermediate reasoning is elicited. Together, these strategies instantiate distinct combinations of the structural dimensions introduced in Section~\ref{sec:preliminary}. We focus on their structural characteristics rather than prompt-level details. Example prompt templates and additional visualizations are provided in the appendix for completeness.

\subsection{Tasks and Models}

To study how reasoning structure interacts with model capacity, we evaluate a diverse set of open-source language models spanning a wide range of parameter scales, from hundreds of millions to over one hundred billion parameters. The selected models cover multiple architectural families and training paradigms, enabling analysis that generalizes beyond any single model class. For clarity, we group models into three capacity regimes: small-scale models with fewer than 5B parameters, medium-scale models between 5B and 15B parameters, and large-scale models exceeding 15B parameters. This categorization allows us to examine how reasoning behavior and failure modes evolve as capacity increases, and to identify thresholds at which more structured reasoning becomes effective.

We evaluate reasoning behavior across multiple task categories that differ in complexity and structure, including mathematical problem solving, logical deduction, commonsense reasoning, and constrained puzzle solving. Concretely, our evaluation includes GSM8K, LogiQA, CommonsenseQA, AIME, and Game24. These tasks are chosen to probe distinct reasoning demands rather than to maximize benchmark coverage.

For adaptive reasoning-control experiments, we focus on \textbf{Qwen-7B-Instruct} as a representative medium-scale model. This setting allows us to directly compare fixed reasoning styles against learned strategy selection while keeping the base model constant.

\subsection{Adaptive Reasoning Control}

Beyond fixed prompting strategies, we study whether a model can learn to \emph{select} an appropriate reasoning style for each input. We consider two selector-based approaches on top of Qwen-7B-Instruct.

We model strategy selection as first predicting a style $s$, then generating an answer conditioned on that choice:
\begin{equation}
s \sim \pi_{\phi}(s \mid x), \qquad y \sim p_{\theta}(y \mid x, s).
\end{equation}
The resulting policy over answers is therefore
\begin{equation}
p(y \mid x) = \sum_{s \in \mathcal{S}} \pi_{\phi}(s \mid x)\, p_{\theta}(y \mid x, s).
\end{equation}
This decomposition separates \emph{reasoning control} from \emph{reasoning execution}: $\pi_{\phi}$ selects how to reason, while $p_{\theta}$ carries out the selected reasoning process.

\paragraph{SFT selector.}
We first train a supervised selector that predicts which reasoning strategy to use for a given problem. The model is trained on task instances paired with strategy labels derived from fixed-style evaluations. Let $s^{*}(x)$ denote the target strategy label for input $x$. The SFT objective is
\begin{equation}
\mathcal{L}_{\mathrm{SFT}}
=
-\mathbb{E}_{x}
\left[
\log \pi_{\phi}\big(s^{*}(x)\mid x\big)
\right].
\end{equation}
At inference time, the selector first outputs a reasoning style, after which the model answers the problem using the corresponding prompting format. This setting tests whether offline supervision is sufficient for learning adaptive reasoning control.

\paragraph{GRPO selector.}
We further study a reinforcement learning variant based on GRPO, where the model learns to select reasoning strategies from outcome-based rewards. For each input $x$, the selector samples a style $s \sim \pi_{\phi}(\cdot \mid x)$, and the base model then produces an answer $y \sim p_{\theta}(y \mid x,s)$. We define the reward as
\begin{equation}
r(x,s,y) = \mathrm{Acc}(y,y^{*}) - \lambda\, C(y) + \gamma\, \mathbf{1}[s \notin \mathcal{S}_{\mathrm{base}} \wedge \hat{y}=y^{*}],
\end{equation}
where $\mathrm{Acc}(y,y^{*})$ is the task reward with respect to the ground-truth answer $y^{*}$, $C(y)$ denotes inference cost measured by token usage, and $\lambda \ge 0$ controls the trade-off between correctness and efficiency. The final term assigns an additional bonus when the model proposes a reasoning style outside the predefined style set $\mathcal{S}_{\mathrm{base}}=\{\mathrm{CoT},\mathrm{CoD},\mathrm{SoT},\mathrm{ToT},\mathrm{AoT}\}$ and still produces a correct answer. This encourages not only effective style selection, but also limited style innovation when the fixed inventory is insufficient. In practice, this bonus encourages the model to try new reasoning styles when the fixed set is not enough.

GRPO optimizes the selector through relative advantages computed within a sampled group of candidate trajectories. At a high level, the objective favors actions whose rewards exceed the group baseline:
\begin{equation}
\max_{\phi}\;
\mathbb{E}_{x}
\left[
\mathbb{E}_{s \sim \pi_{\phi}(\cdot \mid x)}
\big[
A(x,s)\,\log \pi_{\phi}(s\mid x)
\big]
\right],
\end{equation}
where $A(x,s)$ denotes the group-relative advantage induced by sampled rewards. In this way, GRPO provides an online credit-assignment mechanism for reasoning control, in contrast to the imitation-based objective used in SFT. Our goal is not to introduce GRPO as a new reasoning strategy, but to test whether online optimization better captures the trade-offs among existing styles and can move beyond the shallow style collapse observed under supervised training alone.

All experiments are conducted under controlled inference conditions. For fixed-style evaluations, we use single-pass decoding with temperature zero, producing one deterministic output per model and input. This minimizes confounding effects from sampling variance and isolates the effect of reasoning structure. We evaluate model outputs by extracting the final answer and comparing it against the ground-truth label using task-appropriate accuracy metrics. For selector-based methods, we compare SFT and GRPO selectors against the five fixed reasoning styles on the same task set using Qwen-7B-Instruct. This allows us to evaluate both downstream task performance and the quality of the induced reasoning-control policy. Together, these settings support analysis of when structured reasoning improves accuracy, when it increases cost without benefit, and whether reasoning strategy selection can itself be learned adaptively.


\section{Results}
\label{sec:results}

We begin with a high-level summary of aggregate performance across reasoning strategies. For each strategy, we compute mean accuracy over the five tasks, and then average these scores across models within each capacity regime (small, medium, and large). This aggregation reveals broad trends in how reasoning structure interacts with model scale and task difficulty. Figure~\ref{fig:overall_accuracy_barplot} reports the resulting average accuracies for each model group, while detailed per-model and per-task results are provided in Appendix~\ref{appendix: accuracy}.

\subsection{Scaling Effects of Reasoning Structure}

Performance improves with model scale across all reasoning strategies, but the magnitude and nature of this improvement vary substantially. Search-based strategies such as Tree-of-Thought (ToT) and Algorithm-of-Thought (AoT) show stronger scaling effects. They mainly help on harder tasks, including AIME and Game of 24, and only when deployed with large-scale models. On small and medium models, these strategies offer little benefit and fail to outperform simpler alternatives. In contrast, Chain-of-Draft (CoD) demonstrates comparatively stable behavior across model sizes and task types. While it does not consistently achieve the highest absolute accuracy, it avoids severe performance degradation under limited capacity, suggesting that lighter reasoning constraints hold up better when model capacity is limited.
\subsection{When Knowledge Overrides Reasoning}
Results on CommonsenseQA reveal a clear separation between tasks driven primarily by knowledge retrieval and those that benefit from structured reasoning. For large models, all reasoning strategies achieve similarly high accuracy, indicating that performance is largely supported by the model’s internalized knowledge even under minimal reasoning structure. In this regime, the choice of reasoning strategy has little influence on the final outcome.

In contrast, small and medium models perform poorly across all strategies. Even the best-performing configuration, Sketch-of-Thought (SoT) on medium-scale models, achieves accuracy only slightly above 6\%. This is in sharp contrast to large models, where the lowest-performing strategy, Chain-of-Thought (CoT), exceeds 30\% accuracy. The resulting gap suggests that on knowledge-intensive tasks, structured reasoning alone cannot compensate for insufficient model capacity.

\begin{figure}[t]
\centering

\begin{subfigure}[t]{0.49\linewidth}
\centering
\begin{tikzpicture}[
    >=Stealth,
    x=0.82cm,
    y=1.02cm,
    every node/.style={font=\footnotesize}
  ]
  
  \definecolor{searchbased}{RGB}{200,80,80}
  \definecolor{sequential}{RGB}{70,70,180}
  \definecolor{concise}{RGB}{80,160,80}
  \definecolor{regionlight}{RGB}{248,248,248}
  
  \fill[regionlight]    (0.3,-0.1) rectangle (1.9,3.2);
  \fill[regionlight!60] (2.1,-0.1) rectangle (3.9,3.2);
  \fill[regionlight!30] (4.1,-0.1) rectangle (5.9,3.2);
  
  \draw[->, thick] (0,0) -- (6.25,0) node[right, font=\scriptsize] {Model scale};
  \draw[->, thick] (0,0) -- (0,3.45) node[above, font=\scriptsize] {Performance};
  
  \node[below, font=\scriptsize] at (1.1,-0.22) {Small};
  \node[below, font=\scriptsize] at (3.0,-0.22) {Medium};
  \node[below, font=\scriptsize] at (5.0,-0.22) {Large};
  
  \node[font=\scriptsize\bfseries, anchor=west] at (0,3.28) {(a) Reasoning-sensitive tasks};
  
  \draw[searchbased, very thick, smooth] plot coordinates {
    (0.4,0.50) (1.4,0.72) (2.4,1.20) (3.4,1.95) (4.4,2.65) (5.4,3.02)
  };
  \node[searchbased, font=\tiny, fill=white, inner sep=1pt] at (5.05,3.02) {ToT, AoT};
  
  \draw[sequential, thick, smooth] plot coordinates {
    (0.4,0.58) (1.4,1.08) (2.4,1.72) (3.4,2.23) (4.4,2.63) (5.4,2.88)
  };
  \node[sequential, font=\tiny, fill=white, inner sep=1pt] at (3.05,2.36) {CoT};
  
  \draw[concise, thick, dashed, smooth] plot coordinates {
    (0.4,0.64) (1.4,1.14) (2.4,1.76) (3.4,2.26) (4.4,2.67) (5.4,2.92)
  };
  \node[concise, font=\tiny, fill=white, inner sep=1pt] at (1.45,1.78) {CoD, SoT};
  
  \node[
    align=center,
    fill=yellow!18,
    draw=gray!65,
    rounded corners,
    inner sep=2.5pt,
    font=\tiny,
    text width=1.9cm
  ] at (4.85,1.08) {
    \textbf{Finding:}\\
    strategy matters\\
    more at larger scales
  };

\end{tikzpicture}
\end{subfigure}\hfill
\begin{subfigure}[t]{0.49\linewidth}
\centering
\begin{tikzpicture}[
    >=Stealth,
    x=0.82cm,
    y=1.02cm,
    every node/.style={font=\footnotesize}
  ]
  
  \definecolor{knowledge}{RGB}{200,140,60}
  \definecolor{regionlight}{RGB}{248,248,248}
  
  \fill[regionlight]    (0.3,-0.1) rectangle (1.9,3.2);
  \fill[regionlight!60] (2.1,-0.1) rectangle (3.9,3.2);
  \fill[regionlight!30] (4.1,-0.1) rectangle (5.9,3.2);
  
  \draw[->, thick] (0,0) -- (6.25,0) node[right, font=\scriptsize] {Model scale};
  \draw[->, thick] (0,0) -- (0,3.45) node[above, font=\scriptsize] {Performance};
  
  \node[below, font=\scriptsize] at (1.1,-0.22) {Small};
  \node[below, font=\scriptsize] at (3.0,-0.22) {Medium};
  \node[below, font=\scriptsize] at (5.0,-0.22) {Large};
  
  \node[font=\scriptsize\bfseries, anchor=west] at (0,3.28) {(b) Knowledge-intensive tasks};
  
  \draw[knowledge, very thick, smooth] plot coordinates {
    (0.4,0.35) (1.4,0.45) (2.4,0.58) (3.4,1.62) (4.4,2.50) (5.4,2.86)
  };
  \draw[knowledge!80, thick, smooth] plot coordinates {
    (0.4,0.31) (1.4,0.42) (2.4,0.54) (3.4,1.58) (4.4,2.46) (5.4,2.82)
  };
  \draw[knowledge!92, thick, smooth] plot coordinates {
    (0.4,0.34) (1.4,0.48) (2.4,0.60) (3.4,1.65) (4.4,2.54) (5.4,2.89)
  };
  
  \node[knowledge, font=\tiny, fill=white, inner sep=1pt] at (4.85,2.18) {Strategies converge};
  
  \draw[knowledge, very thick, <->, line cap=round] (0.55,0.16) -- (2.45,0.16);
  \node[font=\tiny, above] at (1.50,0.40) {Low sensitivity};
  
  \node[
    align=center,
    fill=yellow!18,
    draw=gray!65,
    rounded corners,
    inner sep=2.5pt,
    font=\tiny,
    text width=1.9cm
  ] at (4.85,1.08) {
    \textbf{Finding:}\\
    capacity dominates,\\
    strategy matters less
  };

\end{tikzpicture}
\end{subfigure}

\caption{Reasoning structure matters more on reasoning-sensitive tasks, while performance converges across strategies on knowledge-intensive tasks.}
\label{fig:reasoning_regimes}
\end{figure}

\subsection{Task Dependence and Strategy Affinity}

Beyond scaling effects, we observe strong task-specific affinities for particular forms of reasoning structure that persist across model sizes.

\textbf{GSM8K.} Despite the availability of  elaborate reasoning strategies, Chain-of-Thought (CoT) consistently achieves the highest accuracy across all capacity regimes. This suggests that for mathematical problems at this level of difficulty, linear stepwise decomposition is sufficient to capture the required reasoning structure. Additional complexity, such as branching or search, does not provide further benefit and may introduce unnecessary overhead.

\textbf{LogiQA.} In contrast, Sketch-of-Thought (SoT) emerges as the most effective strategy, outperforming all alternatives by a substantial margin. Logical reasoning tasks favor concise, structured representations that emphasize relational constraints rather than extended verbal explanations. SoT’s compact, example-driven format appears to guide models toward correct deductions more efficiently, particularly under limited context budgets.

Having characterized task- and scale-dependent behavior under fixed reasoning styles, we next ask whether a model can learn to select among them adaptively. Table~\ref{tab:qwen7b_style_selector_results} compares fixed styles against SFT and GRPO selectors on Qwen-7B-Instruct across all five tasks.

\begin{table}[t]
\centering
\small
\setlength{\tabcolsep}{4.5pt}
\begin{tabular}{lcccccc}
\toprule
\textbf{Method} & \textbf{AIME} & \textbf{CommonsenseQA} & \textbf{GSM8K} & \textbf{Game24} & \textbf{LogiQA} & \textbf{Avg.} \\
\midrule
CoT           & 24.4 & 76.4 & 92.4 & 17.6 & 40.0 & 50.2 \\
CoD           & 25.6 & 81.2 & 92.8 & 22.8 & 47.8 & 54.0 \\
SoT           & 23.4 & 73.0 & 89.8 & 24.4 & 51.8 & 52.5 \\
ToT           & 23.2 & 75.2 & 91.6 & 16.4 & 32.8 & 47.8 \\
AoT           & 23.4 & 79.8 & 92.0 & 15.4 & 29.4 & 48.0 \\
\midrule
SFT selector  & 24.6 & 79.4 & 92.6 & 21.6 & 48.5 & 53.3 \\
GRPO selector & \textbf{27.6} & \textbf{83.4} & \textbf{94.5} & \textbf{34.4} & \textbf{54.2} & \textbf{58.8} \\
\bottomrule
\end{tabular}
\caption{Accuracy (\%) across five reasoning tasks under fixed reasoning styles and learned strategy selectors. GRPO achieves the strongest average performance across all tasks.}
\label{tab:qwen7b_style_selector_results}
\end{table}
\section{Discussion}
\label{sec:discussion}
To move beyond aggregate metrics, we complement our quantitative results with a qualitative analysis of model responses. We use targeted case studies to examine how and why different reasoning strategies succeed or fail under varying task demands and model capacities. This analysis focuses on three questions: (1) what triggers failures for a given strategy on tasks where others succeed, (2) whether the failure reflects a deep problem or a more localized mistake that could be fixed, and (3) what these patterns imply for choosing a reasoning strategy in practice. Together, these case studies reveal recurring failure modes and success patterns that help clarify the mechanisms driving LLM reasoning behavior.

\subsection{Key Findings}

\begin{conclusionbox}
\textbf{Finding 1: Reasoning Reliability Is Capacity-Limited.}  
On challenging tasks, smaller models often fail not because they run out of tokens, but because reasoning breaks down before the computation budget is exhausted. Increased prompting structure cannot compensate for insufficient underlying reasoning capacity.
\end{conclusionbox}

Figure~\ref{fig:token_usage_easy} shows that smaller models do not consistently consume more tokens than larger ones on AIME and Game24, despite substantially lower accuracy. Representative failures in Appendix~\ref{guess-response} further show that these models often terminate with plausible but invalid solutions, suggesting premature commitment rather than insufficient budget. Together, these results indicate that the main limitation of small and medium models is unreliable reasoning, not lack of generation length.

\begin{conclusionbox}
\textbf{Finding 2: Instruction Following Is Also Scale-Dependent.}  
Reliable adherence to output format and reasoning-control instructions improves with model scale. Smaller models frequently violate explicit formatting constraints, reducing robustness even when task understanding is partially correct.
\end{conclusionbox}

As shown by the case studies in the appendix, smaller models often fail to follow simple output requirements such as returning answers in boxed form, and may produce non-canonical answer markers instead. This suggests that structured reasoning depends not only on task-solving ability, but also on fine-grained generation control.

\begin{conclusionbox}
\textbf{Finding 3: Search Improves Coverage, While Concision Improves Efficiency.}  
Search-based strategies are useful on open-ended tasks that require exploration, but they incur higher inference cost. On more structured tasks, concise strategies often achieve comparable accuracy with substantially lower token usage.
\end{conclusionbox}

On Game24, AoT succeeds by exploring and verifying candidate expressions, whereas linear reasoning styles often miss valid solutions and ToT can become inefficient when early branches are poor (Appendix~\ref{sec:aot_comparison}). In contrast, on tasks such as LogiQA, concise strategies such as CoD and SoT reduce token usage substantially while maintaining competitive accuracy (Appendix~\ref{sec:cod_comparison}). This highlights a central trade-off in reasoning design: exploration is valuable when coverage matters, while concise reasoning is preferable when task structure already constrains the solution space.
\subsection{Qualitative Comparison: SFT Selection vs. GRPO Style Synthesis}

To illustrate the qualitative difference between supervised and reinforcement-based reasoning control, we present two representative examples from Qwen-7B-Instruct (see Appendix~\ref{appendix:sft_grpo_comparison} for full response traces). In the first example, the SFT-trained selector chooses one reasoning style from the predefined style set and answers the question accordingly. In the second example, the GRPO-trained model proposes a new hybrid thinking style tailored to the problem, and uses it to solve the problem more effectively. These examples highlight the difference between \emph{style selection} and \emph{style synthesis}.



\section{Conclusion}

Overall, we propose \ourbenchmark, where we studied reasoning structure as a capacity-constrained design choice and used existing strategies as controlled probes to characterize when structure helps and when it does not. Our analyses yield three main findings. First, no single strategy is universally effective: search-based styles benefit open-ended problems at large scale, while concise styles achieve comparable accuracy at much lower cost on structured tasks. Second, failures in small and medium models arise from limited reasoning reliability rather than insufficient computation — these models often terminate prematurely or violate control instructions before exhausting their token budget. Third, SFT for style selection collapses to shallow style preferences, whereas GRPO yields improved adaptive control on Qwen-7B-Instruct, suggesting that online credit assignment is a more promising path than offline supervision. Together, these findings clarify the regimes in which structured reasoning helps, when it is wasteful, and why learning to select a reasoning strategy is itself a challenging inference problem.

\bibliography{colm2026_conference}

@article{sel2023algorithm,
  title={Algorithm of thoughts: Enhancing exploration of ideas in large language models},
  author={Sel, Bilgehan and Al-Tawaha, Ahmad and Khattar, Vanshaj and Jia, Ruoxi and Jin, Ming},
  journal={arXiv preprint arXiv:2308.10379},
  year={2023}
}

@article{yang2025qwen3,
  title={Qwen3 technical report},
  author={Yang, An and Li, Anfeng and Yang, Baosong and Zhang, Beichen and Hui, Binyuan and Zheng, Bo and Yu, Bowen and Gao, Chang and Huang, Chengen and Lv, Chenxu and others},
  journal={arXiv preprint arXiv:2505.09388},
  year={2025}
}

@article{comanici2025gemini,
  title={Gemini 2.5: Pushing the frontier with advanced reasoning, multimodality, long context, and next generation agentic capabilities},
  author={Comanici, Gheorghe and Bieber, Eric and Schaekermann, Mike and Pasupat, Ice and Sachdeva, Noveen and Dhillon, Inderjit and Blistein, Marcel and Ram, Ori and Zhang, Dan and Rosen, Evan and others},
  journal={arXiv preprint arXiv:2507.06261},
  year={2025}
}

@inproceedings{
  guo2025meta,
  title={Meta Thinker: Thinking What {AI} Thinks},
  author={Junyu Guo and Shangding Gu and Costas Spanos and Javad Lavaei},
  booktitle={The 5th Workshop on Mathematical Reasoning and AI at NeurIPS 2025},
  year={2025},
}

@article{chen2024not,
  title={Do not think that much for 2+ 3=? on the overthinking of o1-like LLMs},
  author={Chen, Xingyu and Xu, Jiahao and Liang, Tian and He, Zhiwei and Pang, Jianhui and Yu, Dian and Song, Linfeng and Liu, Qiuzhi and Zhou, Mengfei and Zhang, Zhuosheng and others},
  journal={arXiv preprint arXiv:2412.21187},
  year={2024}
}

@inproceedings{tan2023can,
  title={Can ChatGPT replace traditional KBQA models? An in-depth analysis of the question answering performance of the GPT LLM family},
  author={Tan, Yiming and Min, Dehai and Li, Yu and Li, Wenbo and Hu, Nan and Chen, Yongrui and Qi, Guilin},
  booktitle={International Semantic Web Conference},
  pages={348--367},
  year={2023},
  organization={Springer}
}

@article{fan2024hardmath,
  title={Hardmath: A benchmark dataset for challenging problems in applied mathematics},
  author={Fan, Jingxuan and Martinson, Sarah and Wang, Erik Y and Hausknecht, Kaylie and Brenner, Jonah and Liu, Danxian and Peng, Nianli and Wang, Corey and Brenner, Michael P},
  journal={arXiv preprint arXiv:2410.09988},
  year={2024}
}

@article{zhuo2024bigcodebench,
  title={Bigcodebench: Benchmarking code generation with diverse function calls and complex instructions},
  author={Zhuo, Terry Yue and Vu, Minh Chien and Chim, Jenny and Hu, Han and Yu, Wenhao and Widyasari, Ratnadira and Yusuf, Imam Nur Bani and Zhan, Haolan and He, Junda and Paul, Indraneil and others},
  journal={arXiv preprint arXiv:2406.15877},
  year={2024}
}

@article{wan2025rema,
  title={Rema: Learning to meta-think for LLMs with multi-agent reinforcement learning},
  author={Wan, Ziyu and Li, Yunxiang and Wen, Xiaoyu and Song, Yan and Wang, Hanjing and Yang, Linyi and Schmidt, Mark and Wang, Jun and Zhang, Weinan and Hu, Shuyue and others},
  journal={arXiv preprint arXiv:2503.09501},
  year={2025}
}

@article{austin2021program,
  title={Program Synthesis with Large Language Models},
  author={Austin, Jacob and Odena, Augustus and Nye, Maxwell and Bosma, Maarten and Michalewski, Henryk and Dohan, David and Jiang, Ellen and Cai, Carrie and Terry, Michael and Le, Quoc and others},
  journal={arXiv preprint arXiv:2108.07732},
  year={2021}
}

@article{chen2021evaluating,
  title={Evaluating large language models trained on code},
  author={Chen, Mark and Tworek, Jerry and Jun, Heewoo and Yuan, Qiming and Pinto, Henrique Ponde De Oliveira and Kaplan, Jared and Edwards, Harri and Burda, Yuri and Joseph, Nicholas and Brockman, Greg and others},
  journal={arXiv preprint arXiv:2107.03374},
  year={2021},
}

@article{gao2024meta,
  title={Meta reasoning for large language models},
  author={Gao, Peizhong and Xie, Ao and Mao, Shaoguang and Wu, Wenshan and Xia, Yan and Mi, Haipeng and Wei, Furu},
  journal={arXiv preprint arXiv:2406.11698},
  year={2024}
}

@article{guo2025deepseek,
  title={Deepseek-r1: Incentivizing reasoning capability in LLMs via reinforcement learning},
  author={Guo, Daya and Yang, Dejian and Zhang, Haowei and Song, Junxiao and Zhang, Ruoyu and Xu, Runxin and Zhu, Qihao and Ma, Shirong and Wang, Peiyi and Bi, Xiao and others},
  journal={arXiv preprint arXiv:2501.12948},
  year={2025}
}

@article{yang2025reasonflux,
  title={Reasonflux: Hierarchical LLM reasoning via scaling thought templates},
  author={Yang, Ling and Yu, Zhaochen and Cui, Bin and Wang, Mengdi},
  journal={arXiv preprint arXiv:2502.06772},
  year={2025}
}

@inproceedings{wang2023review,
  title={A review on code generation with LLMs: Application and evaluation},
  author={Wang, Jianxun and Chen, Yixiang},
  booktitle={2023 IEEE International Conference on Medical Artificial Intelligence (MedAI)},
  pages={284--289},
  year={2023},
  organization={IEEE}
}

@article{imani2023mathprompter,
  title={Mathprompter: Mathematical reasoning using large language models},
  author={Imani, Shima and Du, Liang and Shrivastava, Harsh},
  journal={arXiv preprint arXiv:2303.05398},
  year={2023}
}

@article{sui2025stop,
  title={Stop overthinking: A survey on efficient reasoning for large language models},
  author={Sui, Yang and Chuang, Yu-Neng and Wang, Guanchu and Zhang, Jiamu and Zhang, Tianyi and Yuan, Jiayi and Liu, Hongyi and Wen, Andrew and Zhong, Shaochen and Chen, Hanjie and others},
  journal={arXiv preprint arXiv:2503.16419},
  year={2025}
}

@article{fang2025thinkless,
  title={Thinkless: Llm learns when to think},
  author={Fang, Gongfan and Ma, Xinyin and Wang, Xinchao},
  journal={arXiv preprint arXiv:2505.13379},
  year={2025}
}

@article{cobbe2021training,
  title={Training verifiers to solve math word problems},
  author={Cobbe, Karl and Kosaraju, Vineet and Bavarian, Mohammad and Chen, Mark and Jun, Heewoo and Kaiser, Lukasz and Plappert, Matthias and Tworek, Jerry and Hilton, Jacob and Nakano, Reiichiro and others},
  journal={arXiv preprint arXiv:2110.14168},
  year={2021}
}

@article{talmor2018commonsenseqa,
  title={Commonsenseqa: A question answering challenge targeting commonsense knowledge},
  author={Talmor, Alon and Herzig, Jonathan and Lourie, Nicholas and Berant, Jonathan},
  journal={arXiv preprint arXiv:1811.00937},
  year={2018}
}

@article{liu2020logiqa,
  title={Logiqa: A challenge dataset for machine reading comprehension with logical reasoning},
  author={Liu, Jian and Cui, Leyang and Liu, Hanmeng and Huang, Dandan and Wang, Yile and Zhang, Yue},
  journal={arXiv preprint arXiv:2007.08124},
  year={2020}
}

@article{agarwal2025gpt,
  title={gpt-oss-120b \& gpt-oss-20b Model Card},
  author={Agarwal, Sandhini and Ahmad, Lama and Ai, Jason and Altman, Sam and Applebaum, Andy and Arbus, Edwin and Arora, Rahul K and Bai, Yu and Baker, Bowen and Bao, Haiming and others},
  journal={arXiv preprint arXiv:2508.10925},
  year={2025}
}

@inproceedings{shinn2023reflexion,
  title={Reflexion: Language Agents with Verbal Reinforcement Learning},
  author={Shinn, Noah and Cassano, Federico and Labash, Brandon and Gopinath, Avinash and Narasimhan, Karthik and Yao, Shun},
  booktitle={Advances in Neural Information Processing Systems (NeurIPS) Workshops},
  year={2023}
}

@article{yang2024supercorrect,
  title={Supercorrect: Advancing small llm reasoning with thought template distillation and self-correction},
  author={Yang, Ling and Yu, Zhaochen and Zhang, Tianjun and Xu, Minkai and Gonzalez, Joseph E and Cui, Bin and Yan, Shuicheng},
  journal={arXiv preprint arXiv:2410.09008},
  year={2024}
}

@article{meng2024llm,
  title={Llm-a*: Large language model enhanced incremental heuristic search on path planning},
  author={Meng, Silin and Wang, Yiwei and Yang, Cheng-Fu and Peng, Nanyun and Chang, Kai-Wei},
  journal={arXiv preprint arXiv:2407.02511},
  year={2024}
}

@article{chen2024expanding,
  title={Expanding performance boundaries of open-source multimodal models with model, data, and test-time scaling},
  author={Chen, Zhe and Wang, Weiyun and Cao, Yue and Liu, Yangzhou and Gao, Zhangwei and Cui, Erfei and Zhu, Jinguo and Ye, Shenglong and Tian, Hao and Liu, Zhaoyang and others},
  journal={arXiv preprint arXiv:2412.05271},
  year={2024}
}

@article{snell2024scaling,
  title={Scaling llm test-time compute optimally can be more effective than scaling model parameters},
  author={Snell, Charlie and Lee, Jaehoon and Xu, Kelvin and Kumar, Aviral},
  journal={arXiv preprint arXiv:2408.03314},
  year={2024}
}

@article{balachandran2025inference,
  title   = {Inference-Time Scaling for Complex Tasks: Where We Stand and What Lies Ahead},
  author  = {Balachandran, Vidhisha and Chen, Jingya and Chen, Lingjiao and Garg, Shivam and Joshi, Neel and Lara, Yash and Langford, John and Nushi, Besmira and Vineet, Vibhav and Wu, Yue and Yousefi, Safoora},
  journal = {arXiv preprint arXiv:2504.00294},
  year    = {2025}
}

@article{bi2024forest,
  title   = {Forest-of-Thought: Scaling Test-Time Compute for Enhancing {LLM} Reasoning},
  author  = {Bi, Zhenni and Han, Kai and Liu, Chuanjian and Tang, Yehui and Wang, Yunhe},
  journal = {arXiv preprint arXiv:2412.09078},
  year    = {2024}
}

@article{metaReasoner2025,
  title   = {Meta-Reasoner: Dynamic Guidance for Optimized Inference-time Reasoning in Large Language Models},
  author  = {Sui, Yuan and He, Yufei and Cao, Tri and Han, Simeng and Chen, Yulin and Hooi, Bryan},
  journal = {arXiv preprint arXiv:2502.19918},
  year    = {2025}
}

@article{zhu2025agents,
  title   = {Scaling Test-time Compute for {LLM} Agents},
  author  = {Zhu, King and Li, Hanhao and Wu, Siwei and Xing, Tianshun and Ma, Dehua and Tang, Xiangru and Liu, Minghao and Yang, Jian and Liu, Jiaheng and Jiang, Yuchen Eleanor and Zhang, Changwang and Lin, Chenghua and Wang, Jun and Zhang, Ge and Zhou, Wangchunshu},
  journal = {arXiv preprint arXiv:2506.12928},
  year    = {2025}
}

@article{wang2025nowait,
  title   = {Wait, We Don{\textquoteright}t Need to ``Wait''! Removing Thinking Tokens Improves Reasoning Efficiency},
  author  = {Wang, Chenlong and Feng, Yuanning and Chen, Dongping and Chu, Zhaoyang and Krishna, Ranjay and Zhou, Tianyi},
  journal = {arXiv preprint arXiv:2506.08343},
  year    = {2025}
}

@article{wu2025moreless,
  title   = {When More is Less: Understanding Chain-of-Thought Length in {LLMs}},
  author  = {Wu, Yuyang and Wang, Yifei and Du, Tianqi and Jegelka, Stefanie and Wang, Yisen},
  journal = {arXiv preprint arXiv:2502.07266},
  year    = {2025}
}

@article{besta2023graph,
  title   = {Graph of Thoughts: Solving Elaborate Problems with Large Language Models},
  author  = {Besta, Maciej and Blach, Nils and Kubicek, Ales and Gerstenberger, Robert and Podstawski, Micha{\l} and Gianinazzi, Lukas and Gajda, Joanna and Lehmann, Tomasz and Niewiadomski, Hubert and Nyczyk, Piotr and Hoefler, Torsten},
  journal = {arXiv preprint arXiv:2308.09687},
  year    = {2023}
}

@article{yao2023graphthoughts,
  title   = {Beyond Chain-of-Thought, Effective Graph-Of-Thought Reasoning in Large Language Models},
  author  = {Yao, Yao and Yu, Li and Zhang, Hongming and Liu, Zhenghao and Han, Xianpei and Sun, Le},
  journal = {arXiv preprint arXiv:2305.16582},
  year    = {2023}
}

@article{renze2024self,
  title={Self-reflection in llm agents: Effects on problem-solving performance},
  author={Renze, Matthew and Guven, Erhan},
  journal={arXiv preprint arXiv:2405.06682},
  year={2024}
}

@article{aytes2025sketch,
  title={Sketch-of-thought: Efficient LLM reasoning with adaptive cognitive-inspired sketching},
  author={Aytes, Simon A and Baek, Jinheon and Hwang, Sung Ju},
  journal={arXiv preprint arXiv:2503.05179},
  year={2025}
}

@article{xu2025chain,
  title={Chain of draft: Thinking faster by writing less},
  author={Xu, Silei and Xie, Wenhao and Zhao, Lingxiao and He, Pengcheng},
  journal={arXiv preprint arXiv:2502.18600},
  year={2025}
}

@article{yasunaga2023large,
  title={Large language models as analogical reasoners},
  author={Yasunaga, Michihiro and Chen, Xinyun and Li, Yujia and Pasupat, Panupong and Leskovec, Jure and Liang, Percy and Chi, Ed H and Zhou, Denny},
  journal={arXiv preprint arXiv:2310.01714},
  year={2023}
}

@article{wei2022chain,
  title={Chain-of-thought prompting elicits reasoning in large language models},
  author={Wei, Jason and Wang, Xuezhi and Schuurmans, Dale and Bosma, Maarten and Xia, Fei and Chi, Ed and Le, Quoc V and Zhou, Denny and others},
  journal={Advances in neural information processing systems},
  volume={35},
  pages={24824--24837},
  year={2022}
}

@article{huang2025gemini,
  title={Gemini 2.5 Pro capable of winning gold at IMO 2025},
  author={Huang, Yichen and Yang, Lin F},
  journal={arXiv preprint arXiv:2507.15855},
  year={2025}
}

@article{yao2023tree,
  title={Tree of thoughts: Deliberate problem solving with large language models},
  author={Yao, Shunyu and Yu, Dian and Zhao, Jeffrey and Shafran, Izhak and Griffiths, Tom and Cao, Yuan and Narasimhan, Karthik},
  journal={Advances in neural information processing systems},
  volume={36},
  pages={11809--11822},
  year={2023}
}
\bibliographystyle{colm2026_conference}

\appendix
\raggedbottom
\section{Related Work}

\subsection{Reasoning with Large Language Models}
Recent advances in LLM reasoning have been driven by the development of structured \emph{thinking styles}. Chain-of-Thought (CoT) prompting~\citep{wei2022chain} demonstrated that step-by-step reasoning can substantially improve performance, particularly in mathematical and logical tasks. Building on this idea, Tree-of-Thought (ToT)~\citep{yao2023tree} introduced a branching exploration strategy that allows models to consider multiple reasoning paths in parallel. Subsequent paradigms such as Algorithm-of-Thought (AoT)~\citep{sel2023algorithm}, Sketch-of-Thought (SoT)~\citep{aytes2025sketch}, and Chain-of-Draft (CoD)~\citep{xu2025chain} further extend this direction by incorporating algorithmic priors, lightweight reasoning sketches, or iterative drafting mechanisms into prompts. Other approaches leverage high-level templates or rich contextual information to equip LLMs with more structured reasoning capabilities~\citep{gao2024meta,yasunaga2023large,yang2025reasonflux,yang2024supercorrect}. Despite these advances, most existing methods rely on a fixed reasoning style determined in advance, which may be suboptimal across heterogeneous tasks.

\subsection{Benchmarking LLM Reasoning}

The development of comprehensive benchmarks has been crucial for evaluating the reasoning capabilities of Large Language Models (LLMs). Existing research has largely focused on specialized domains, each requiring distinct reasoning skills. Mathematical reasoning is commonly assessed using benchmarks such as GSM8K~\citep{cobbe2021training}, HardMath~\citep{fan2024hardmath} and the more challenging AIME problems. For logical reasoning, datasets like LogiQA~\citep{liu2020logiqa} provide standardized tests, while commonsense reasoning is typically measured by benchmarks such as CommonsenseQA~\citep{talmor2018commonsenseqa}. In the domain of code generation, Bigcodebench~\citep{zhuo2024bigcodebench}, HumanEval~\citep{chen2021evaluating} and MBPP~\citep{austin2021program} evaluate functional correctness and algorithmic problem-solving capabilities. More recently, many works have focused on assessing LLMs' performance on  puzzle-solving and constrained reasoning tasks. These include problems such as word sorting, Sudoku, and Game of 24, which require structured, multi-step deduction and explore the limits of LLMs' systematic reasoning abilities.

\subsection{Test-Time Reasoning}
A growing body of work studies how to \emph{allocate, structure, and regulate} computation during inference, with the goal of improving reasoning without changing model weights. Test time Scaling serves as a promising direction~\cite{chen2024expanding,snell2024scaling}, and recent frameworks also  show that increasing test time compute via repeated sampling, search, self correction, or ensemble style aggregation can yield notable gains, but the benefits are highly task dependent and often saturate on harder regimes \citep{balachandran2025inference,bi2024forest}. Beyond generic scaling, structured exploration methods extend linear scratchpads into explicit search spaces, including tree or graph based reasoning and agentic reflection loops \citep{yao2023graphthoughts,besta2023graph,shinn2023reflexion,zhu2025agents,meng2024llm}. At the same time, multiple recent studies highlight an \emph{overthinking} failure mode, where longer reasoning traces increase cost and may even reduce accuracy, motivating adaptive halting and inference time control policies \citep{wu2025moreless,wang2025nowait,metaReasoner2025}. These results suggest that test time reasoning is not a monotone knob: its effectiveness depends on the interaction between task structure, model capacity, and the form of reasoning control. Our work complements this line by empirically characterizing when added structure helps, when it is redundant, and when it induces systematic breakdowns that look like reasoning but do not function as reasoning.

\section{Sample Responses}
\label{appendix:sample_response}

\subsection{Qualitative Comparison: SFT Selection vs. GRPO Style Synthesis}
\label{appendix:sft_grpo_comparison}

To illustrate the qualitative difference between supervised and reinforcement-based reasoning control, we present two representative examples from Qwen-7B-Instruct. In the first example, the SFT-trained selector chooses one reasoning style from the predefined style set and answers the question accordingly. In the second example, the GRPO-trained model proposes a new hybrid thinking style tailored to the problem and uses it to solve the problem more effectively. These examples highlight the difference between \emph{style selection} and \emph{style synthesis}.

\begin{tcolorbox}[
    enhanced,
    breakable,
    colback=blue!3,
    colframe=blue!55!black,
    boxrule=0.5pt,
    arc=2pt,
    left=5pt,right=5pt,top=4pt,bottom=4pt,
    before skip=4pt,
    after skip=4pt,
    boxsep=1pt,
    title=\textbf{Example 1: SFT-Trained Selector Chooses a Predefined Style},
    fonttitle=\bfseries
]
\textbf{Problem.}
A store sells notebooks in packs of $6$ and pens in packs of $4$. A student buys some packs of each and ends up with exactly $34$ items in total. What is the smallest possible number of packs the student could have bought?

\textbf{Predicted style.}
\textit{Chain-of-Thought (CoT)}

\textbf{Model response.}
We want nonnegative integers $x$ and $y$ such that
\[
6x+4y=34.
\]
Dividing by $2$ gives
\[
3x+2y=17.
\]
Now try values of $x$ such that $17-3x$ is nonnegative and even:
\begin{itemize}
    \item If $x=1$, then $2y=14$, so $y=7$ and $x+y=8$.
    \item If $x=3$, then $2y=8$, so $y=4$ and $x+y=7$.
    \item If $x=5$, then $2y=2$, so $y=1$ and $x+y=6$.
\end{itemize}
Thus the smallest valid total is $6$.

\textbf{Final answer.}
\[
\boxed{6}
\]

\textbf{Observation.}
The SFT-trained selector successfully chooses a valid reasoning style from the predefined inventory and applies it correctly. However, the control policy remains confined to selecting among existing styles rather than adapting the reasoning format itself.
\end{tcolorbox}

\begin{tcolorbox}[
    enhanced,
    breakable,
    colback=orange!4,
    colframe=orange!70!black,
    boxrule=0.5pt,
    arc=2pt,
    left=5pt,right=5pt,top=4pt,bottom=4pt,
    before skip=4pt,
    after skip=4pt,
    boxsep=1pt,
    title=\textbf{Example 2: GRPO-Trained Model Synthesizes a New Style on Game24},
    fonttitle=\bfseries
]
\textbf{Problem.}
Use the numbers $4$, $7$, $8$, and $8$ exactly once, together with arithmetic operations, to obtain $24$.

\textbf{Proposed style.}
\textit{Pair-and-Target Search}

\textbf{Style rationale.}
Instead of following a purely linear chain or a broad unconstrained search, first form promising local pairs, then match them to a target value that can be combined into $24$. This style mixes lightweight search with explicit target checking.

\textbf{Model response.}
I will use a \emph{Pair-and-Target Search} strategy:
\begin{enumerate}
    \item Look for simple pairings among the numbers.
    \item Check whether the remaining numbers can transform the intermediate result into $24$.
    \item Prefer pairings that create easy target values such as $1$, $2$, $3$, $4$, $6$, $8$, $12$, or $24$.
\end{enumerate}

Pair the two $8$'s:
\[
8 \div 8 = 1.
\]
Now the remaining numbers are $4$ and $7$, together with the intermediate value $1$. Next combine $7$ and $1$:
\[
7 - 1 = 6.
\]
Then multiply by the remaining $4$:
\[
4 \times 6 = 24.
\]
So a valid expression is
\[
4 \times \left(7 - \frac{8}{8}\right)=24.
\]

\textbf{Final answer.}
\[
\boxed{4 \times \left(7 - \frac{8}{8}\right)=24}
\]

\textbf{Observation.}
Unlike the SFT-trained selector, the GRPO-trained model does not merely choose from the predefined style set. Instead, it synthesizes a task-specific hybrid reasoning pattern that combines local search with target-based verification.
\end{tcolorbox}

\textbf{Note on out-of-set style frequency.}
The GRPO reward includes a bonus term for proposing a reasoning style outside $\mathcal{S}_{\mathrm{base}} = \{\mathrm{CoT}, \mathrm{CoD}, \mathrm{SoT}, \mathrm{ToT}, \mathrm{AoT}\}$ when the answer is still correct (see Section~3). In our experiments on Qwen-7B-Instruct, out-of-set styles were proposed in approximately 12\% of Game24 trajectories and less than 3\% of trajectories on the remaining four tasks, consistent with the intuition that style synthesis is most useful on open-ended combinatorial problems where no predefined style is well-matched. The example above (\textit{Pair-and-Target Search}) is representative of the kinds of hybrid strategies the model constructs in this regime. On structured tasks such as GSM8K and LogiQA, the model predominantly selects from the predefined set, using the bonus as a safety valve rather than a primary mechanism.

\subsection{Incorrect Response Examples}
\label{sec:small_model_response}

This response was produced by the Qwen2-3B model.

\begin{llmresponse}\label{guess-response}
\textbf{Question:} The pages of a book are numbered $1$ through $n$. When the page numbers were added, one page number was mistakenly added twice, resulting in an incorrect sum of $1986$. What was the page number that was added twice?

\textbf{Ground Truth:} $33$

\textbf{LLM Response:}

Here is a step-by-step solution.

\textbf{1. Formula for the sum of page numbers.}
\[
S = \frac{n(n+1)}{2}
\]

\textbf{2. Set up the equation.}
Let $x$ be the page number added twice. Then the incorrect sum is
\[
\frac{n(n+1)}{2}+x=1986.
\]

\textbf{3. Trial and error.}
If $n=60$, then
\[
\frac{60\cdot 61}{2}=1830.
\]
So
\[
x = 1986-1830=156.
\]

\textbf{4. Verification.}
The model concludes that $156$ could be the duplicated page number.

\[
\boxed{156}
\]
\end{llmresponse}

This response was produced by Llama3-8B on the Game24 task.

\begin{game24response}\label{box:game24-puzzle}
\textbf{Puzzle:} Use the numbers $3$, $3$, $12$, and $12$ exactly once each with operations $+$, $-$, $\times$, $\div$ and parentheses to make $24$.

\textbf{Ground Truth:} One valid solution is
\[
(12 \div 3)\times(3+3)=24.
\]

\textbf{LLM Response:}

Here is one attempted solution.

\textbf{1. Look for obvious pairs.}
\begin{itemize}
    \item $12 \times 12 = 144$, which is too large.
    \item $3 \times 12 = 36$, which is also too large.
\end{itemize}

\textbf{2. Consider division.}
\[
12 \div 3 = 4.
\]

\textbf{3. Build from there.}
Use $4 \times 3 = 12$.

\textbf{4. Combine the results.}
\[
12 + 12 = 24.
\]

\textbf{Therefore, the solution is}
\[
\boxed{12 + 12}
\]
\end{game24response}
\subsection{Comparison Between Different Styles}
\label{sec:aot_comparison}

\textbf{Problem.} Use the numbers $2$, $3$, $13$, and $13$ exactly once each with operations $+$, $-$, $\times$, $\div$ and parentheses to make $24$.

\textbf{Ground Truth.} A valid solution is $(13 \times 3 - 13) - 2 = 24$.

\begin{tcolorbox}[title=\textbf{CoT / ToT / CoD / SoT Summary}, breakable, enhanced]
\textbf{CoT.} Produces multiple incorrect trials and concludes that no solution exists.

\textbf{ToT.} Explores several candidate branches but still concludes incorrectly that there is no solution.

\textbf{CoD.} Generates short draft attempts but also ends with the incorrect conclusion that there is no solution.

\textbf{SoT.} Begins systematic enumeration but is truncated before reaching a valid final answer.
\end{tcolorbox}

\begin{tcolorbox}[title=\textbf{AoT Response (Correct)}, breakable, enhanced]
\textbf{Forward analysis.}
The model identifies $13 \times 3 = 39$ as a promising intermediate result, then reduces:
\[
39 - 13 = 26,\qquad 26 - 2 = 24.
\]

\textbf{Verification.}
\[
(13 \times 3 - 13) - 2 = 24.
\]

\textbf{Conclusion.}
AoT succeeds because it combines directed search with explicit verification, whereas the other styles either terminate too early or fail to cover the relevant solution path.
\end{tcolorbox}

\subsection{Reasoning Style Comparison: Conciseness}
\label{sec:cod_comparison}

\begin{commonsenseqa}
\label{box:commonsense-qa-styles}

\textbf{Problem:} Reading newspaper is one of many ways to practice your what?

\textbf{Options:} A) literacy \quad B) knowing how to read \quad C) money \quad D) buying \quad E) money bank

\textbf{Ground Truth:} A) literacy

\textbf{Chain of Thought (CoT) Response.} \textit{[Length: 1,192 characters]}
The model gives a step-by-step explanation, compares all options, and concludes that reading a newspaper helps practice literacy.
\[
\boxed{A}
\]

\textbf{Tree of Thoughts (ToT) Response.} \textit{[Length: 1,912 characters]}
The model simulates a multi-expert discussion, eliminates irrelevant options, and selects literacy as the best answer.
\[
\boxed{A}
\]

\textbf{Analysis of Thoughts (AoT) Response.} \textit{[Length: 1,779 characters]}
The model performs a forward analysis of the options and a brief verification stage before selecting the correct answer.
\[
\boxed{A}
\]

\textcolor{blue}{\textbf{Chain of Draft (CoD) Response.}} \textit{\textcolor{blue}{[Length: 1,005 characters, 16\% shorter]}}
\textcolor{blue}{The model gives a shorter but still explicit comparison of the options and concludes with the correct answer.}
\[
\textcolor{blue}{\boxed{A}}
\]

\textcolor{green}{\textbf{Sketch of Thought (SoT) Response.}} \textit{\textcolor{green}{[Length: 69 characters, 94\% shorter]}}
\begin{center}
\textcolor{green}{\texttt{\#reading\_newspaper -> \#improves -> \#literacy}}
\end{center}
\[
\textcolor{green}{\boxed{A}}
\]
\end{commonsenseqa}

\subsection{Token Usage}

\begin{figure*}[htbp]
    \centering
    \begin{subfigure}[t]{0.48\textwidth}
        \centering
        \includegraphics[width=\linewidth]{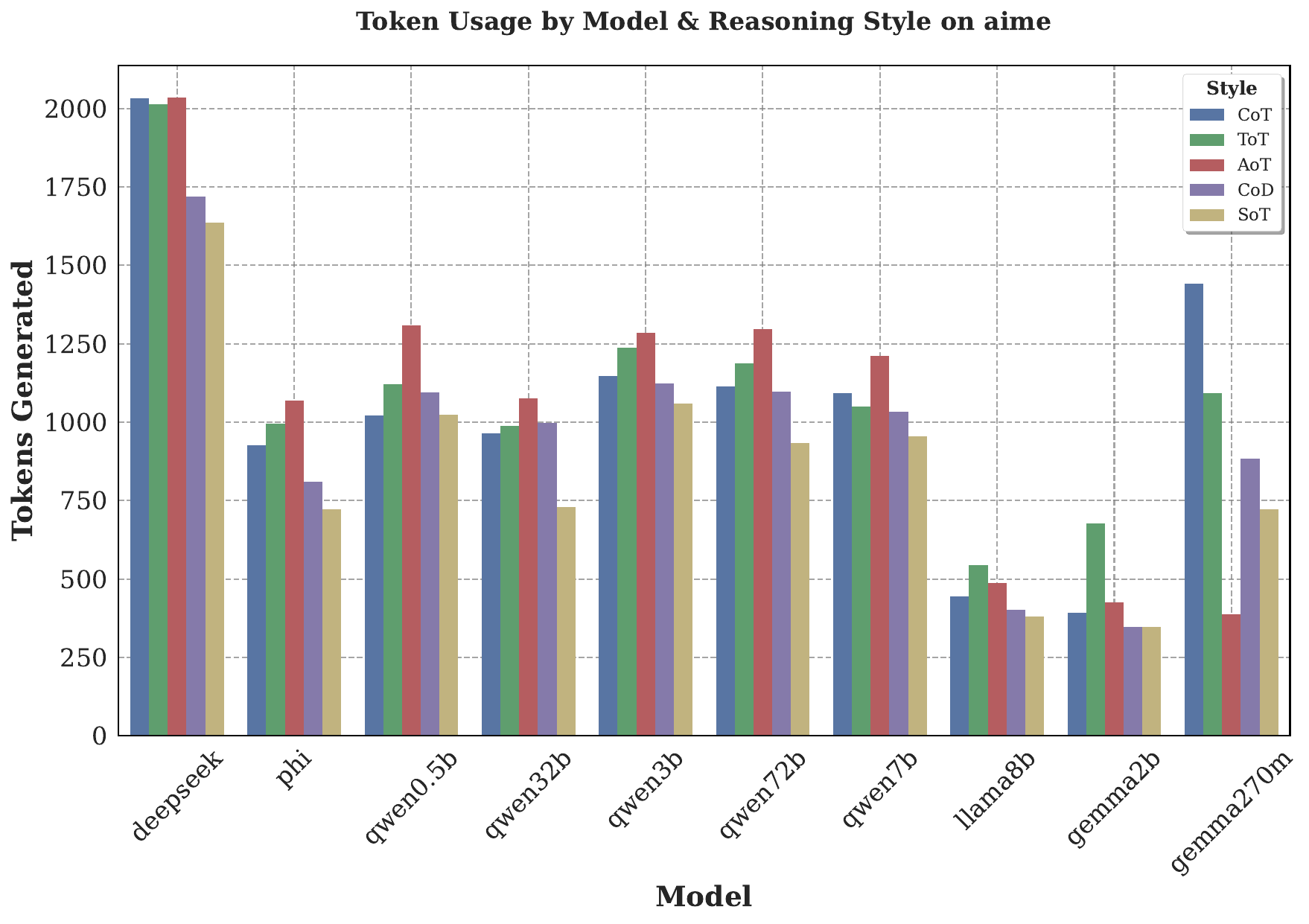}
        \caption{AIME token usage across all models and reasoning styles.}
        \label{fig:aime_token_usage}
    \end{subfigure}
    \hfill
    \begin{subfigure}[t]{0.48\textwidth}
        \centering
        \includegraphics[width=\linewidth]{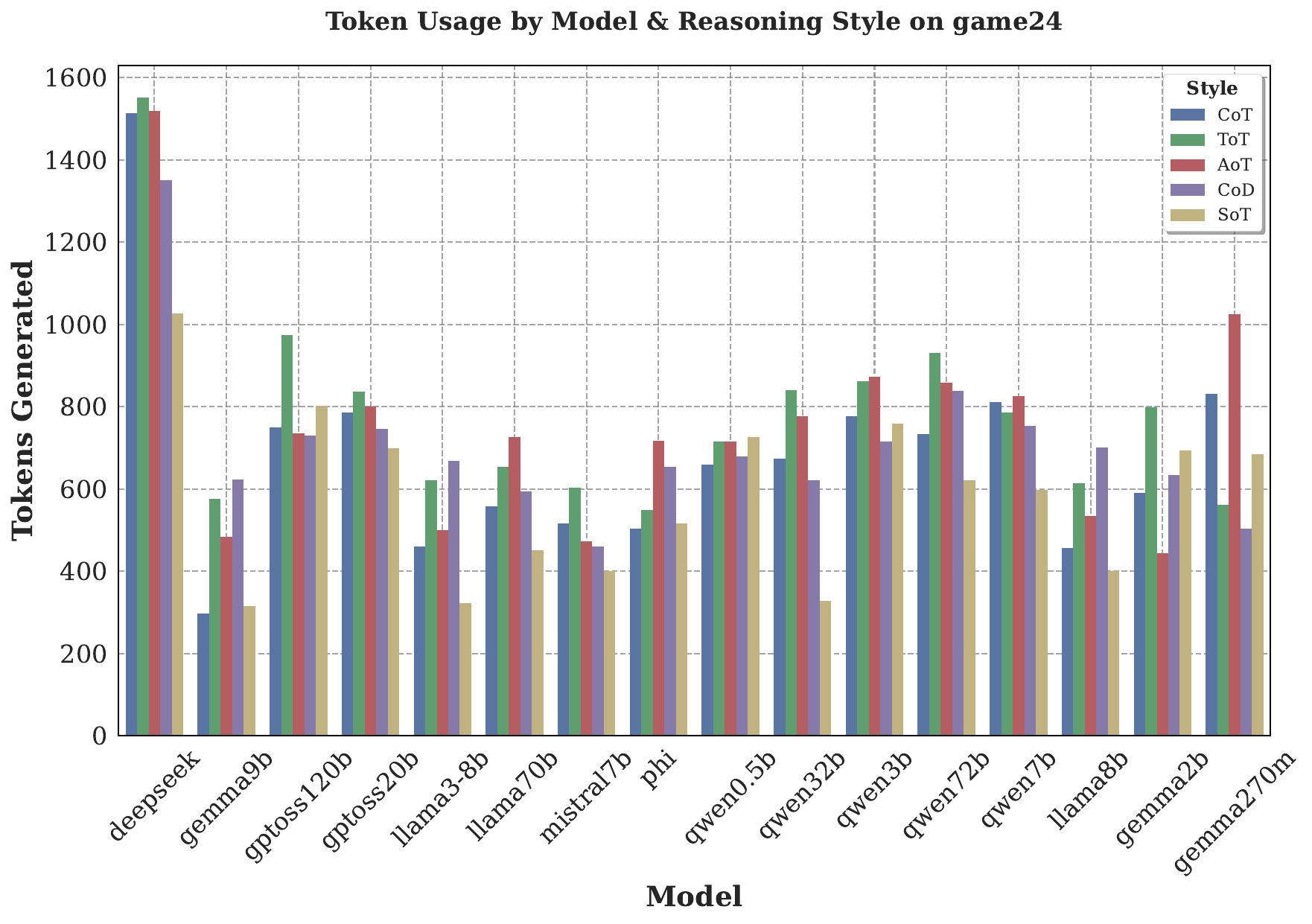}
        \caption{Game24 token usage across all models and reasoning styles.}
        \label{fig:game24_token_usage}
    \end{subfigure}
    \caption{Token consumption patterns across models and reasoning styles for AIME and Game24. Search-based strategies consume more tokens and exhibit higher variance across model scales, while concise strategies maintain lower and more stable token budgets.}
    \label{fig:token_usage_hard}
\end{figure*}

\begin{figure*}[htbp]
    \centering
    \begin{subfigure}[t]{0.48\textwidth}
        \centering
        \includegraphics[width=\linewidth]{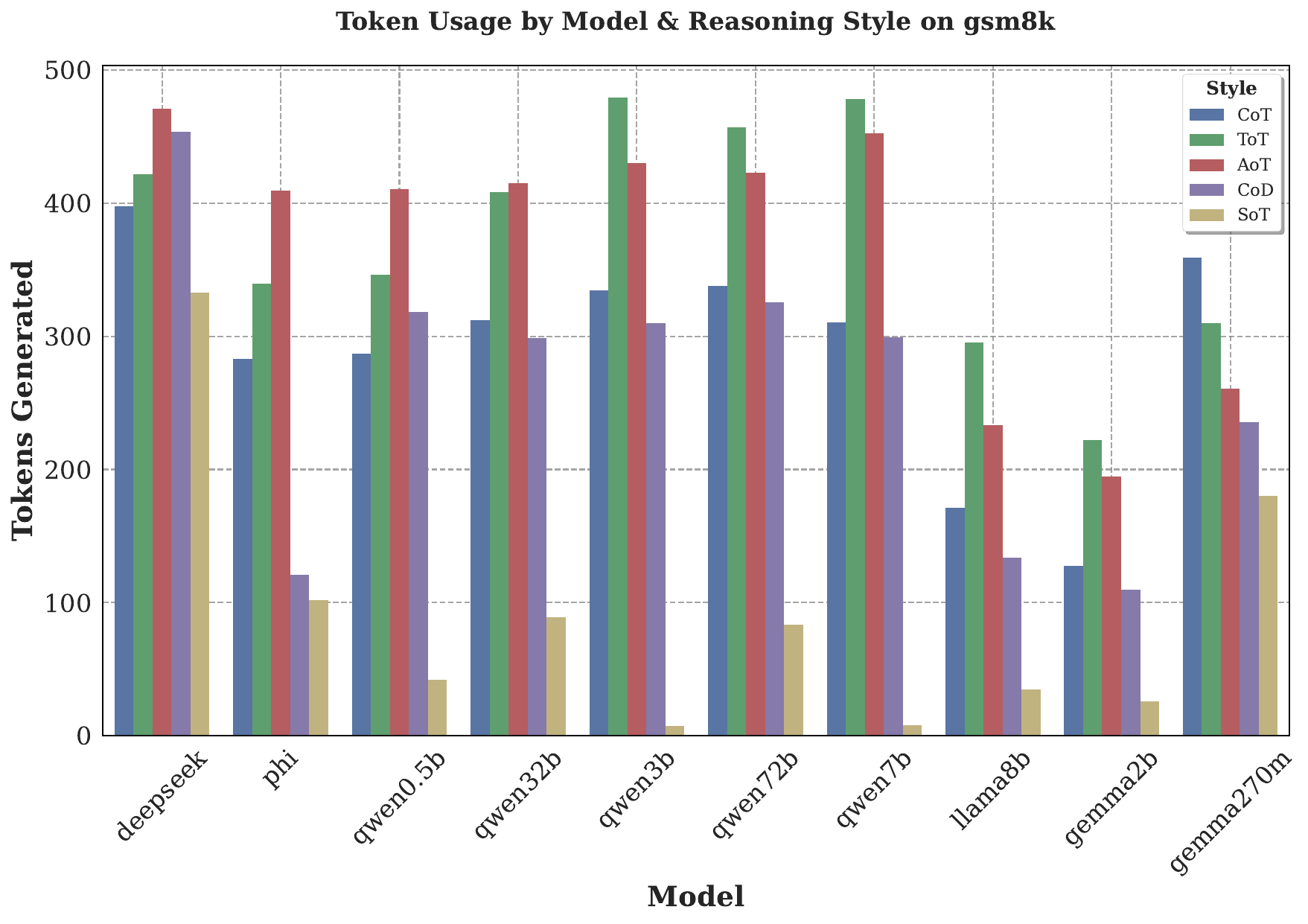}
        \caption{GSM8K token usage across all models and reasoning styles.}
        \label{fig:gsm8k_token_usage}
    \end{subfigure}
    \hfill
    \begin{subfigure}[t]{0.48\textwidth}
        \centering
        \includegraphics[width=\linewidth]{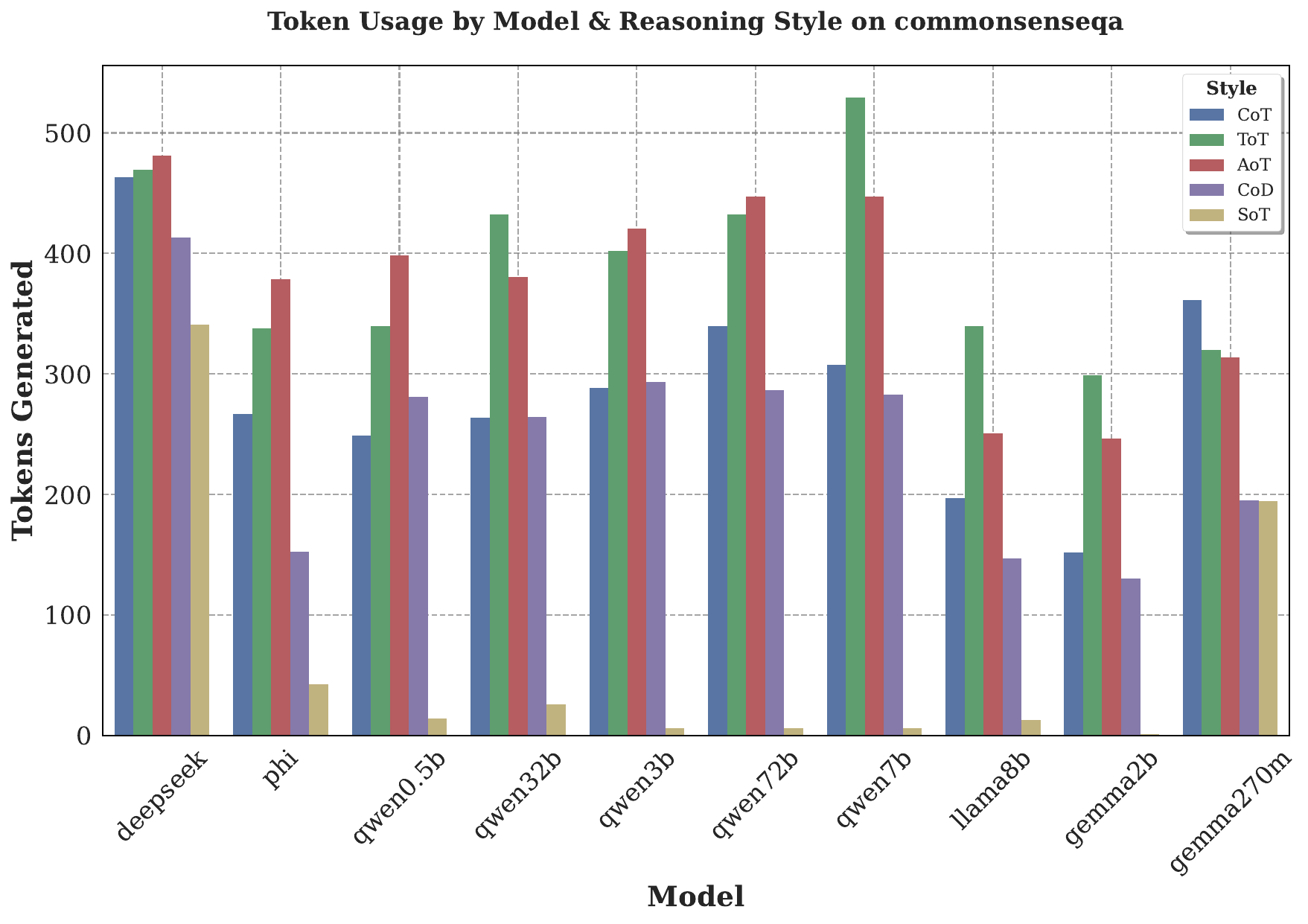}
        \caption{CommonsenseQA token usage across all models and reasoning styles.}
        \label{fig:commonsenseqa_token_usage}
    \end{subfigure}
    \caption{Token consumption patterns across models and reasoning styles for GSM8K and CommonsenseQA. Concise strategies remain substantially cheaper, while heavier reasoning styles often incur additional cost with smaller gains on knowledge-intensive tasks.}
    \label{fig:token_usage_easy}
\end{figure*}
\section{Thinking Styles}
\label{appendix: thinking styles}
\begin{figure}[htbp]
    \centering
    \includegraphics[scale=0.7]{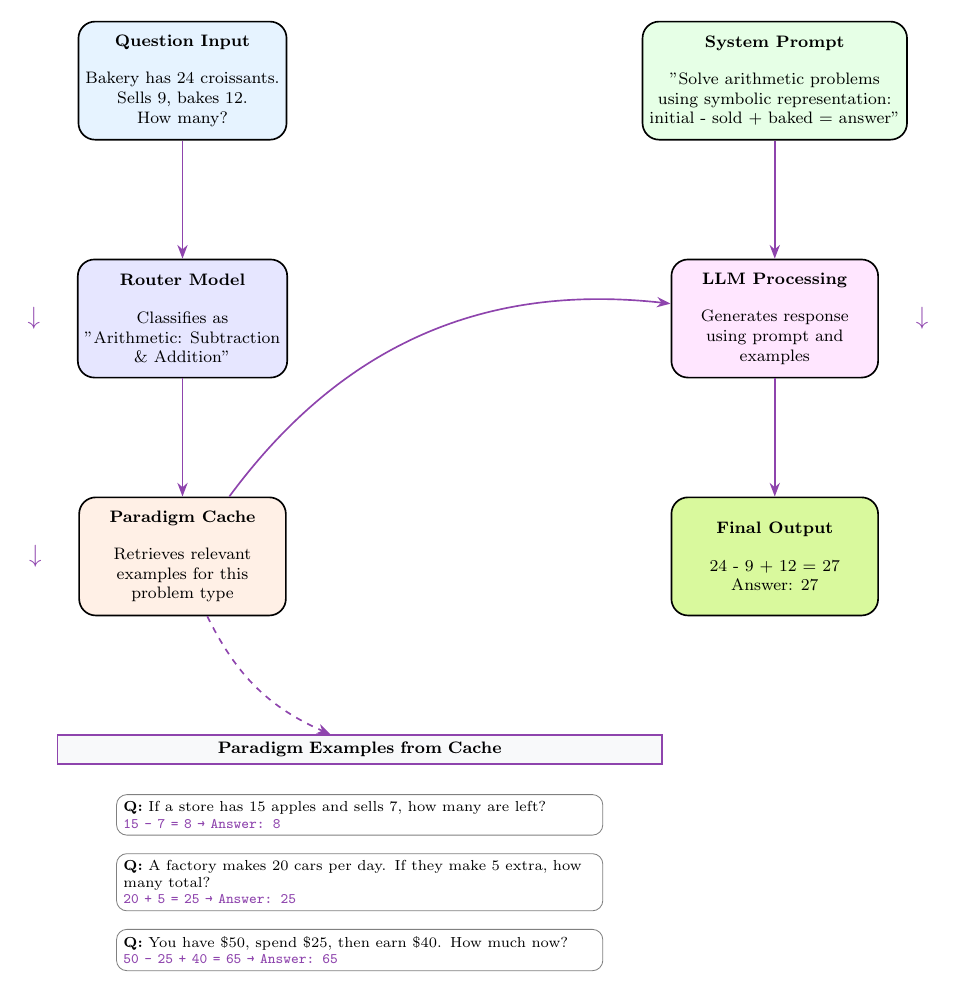}
    \caption{Sketch-of-Thought (SoT): Router-based paradigm selection with exemplar retrieval. The method classifies the input problem, retrieves relevant examples from a paradigm cache, applies targeted prompts, and generates responses through structured LLM processing.}
    \label{fig:sot}
\end{figure}
\begin{figure}[htbp]
    \centering
    \begin{subfigure}[b]{\textwidth}
        \centering
        \includegraphics[width=0.3\textwidth]{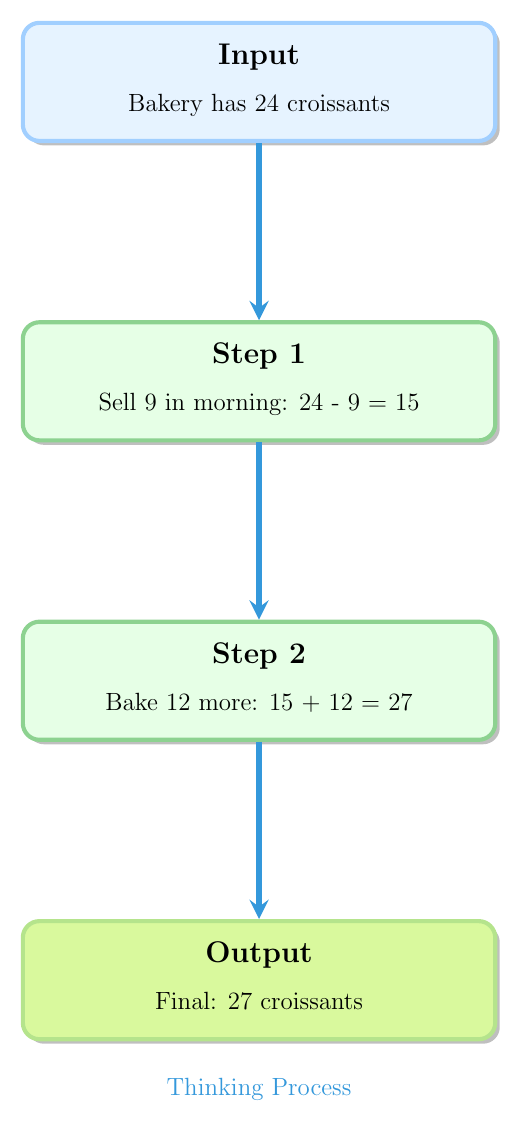}
        \caption{Chain-of-Thought (CoT): Linear step-by-step reasoning}
        \label{fig:cot}
    \end{subfigure}
    
    \vspace{0.8cm}
    
    \begin{subfigure}[b]{\textwidth}
        \centering
        \includegraphics[width=0.5\textwidth]{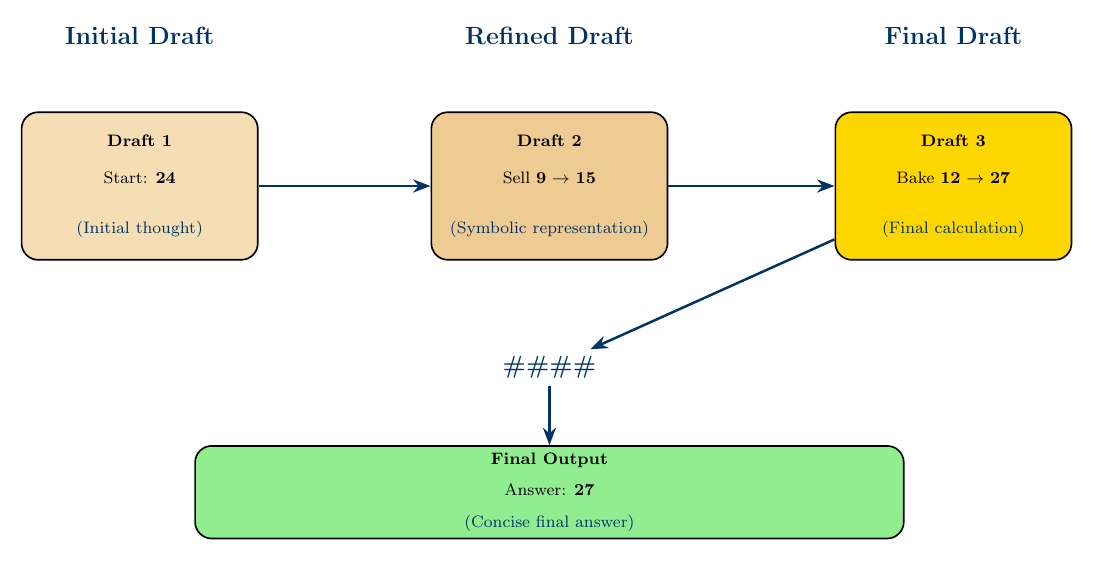}
        \caption{Chain-of-Draft (CoD): Iterative refinement process}
        \label{fig:cod}
    \end{subfigure}
    \caption{Sequential reasoning methodologies: CoT follows a linear progression while CoD employs iterative refinement of drafts.}
    \label{fig:sequential_methods}
\end{figure}

\begin{figure}[htbp]
    \centering
    \begin{subfigure}[b]{\textwidth}
        \centering
        \includegraphics[width=0.9\textwidth]{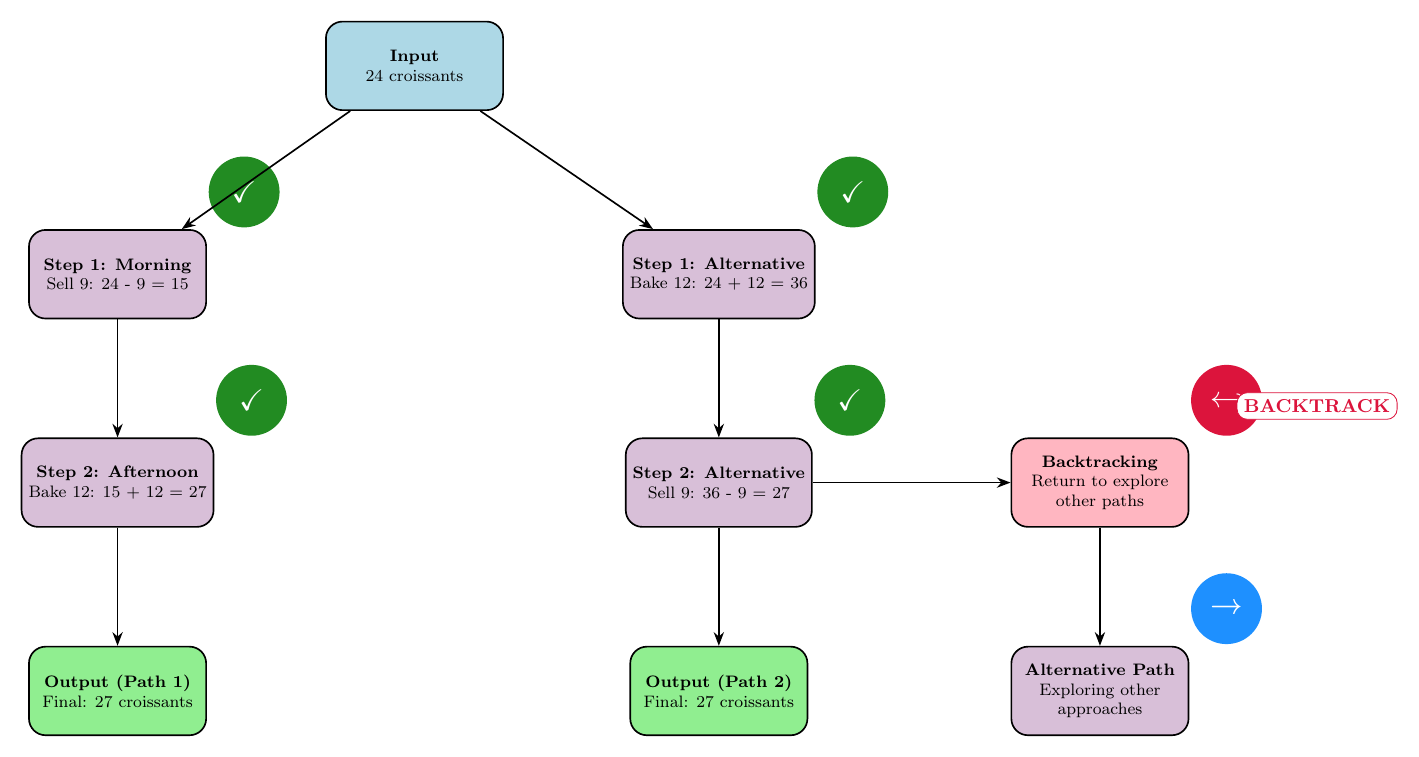}
        \caption{Algorithm-of-Thought (AoT): Backtracking exploration}
        \label{fig:tot}
    \end{subfigure}
    
    \vspace{0.8cm}
    
    \begin{subfigure}[b]{\textwidth}
        \centering
        \includegraphics[width=0.9\textwidth]{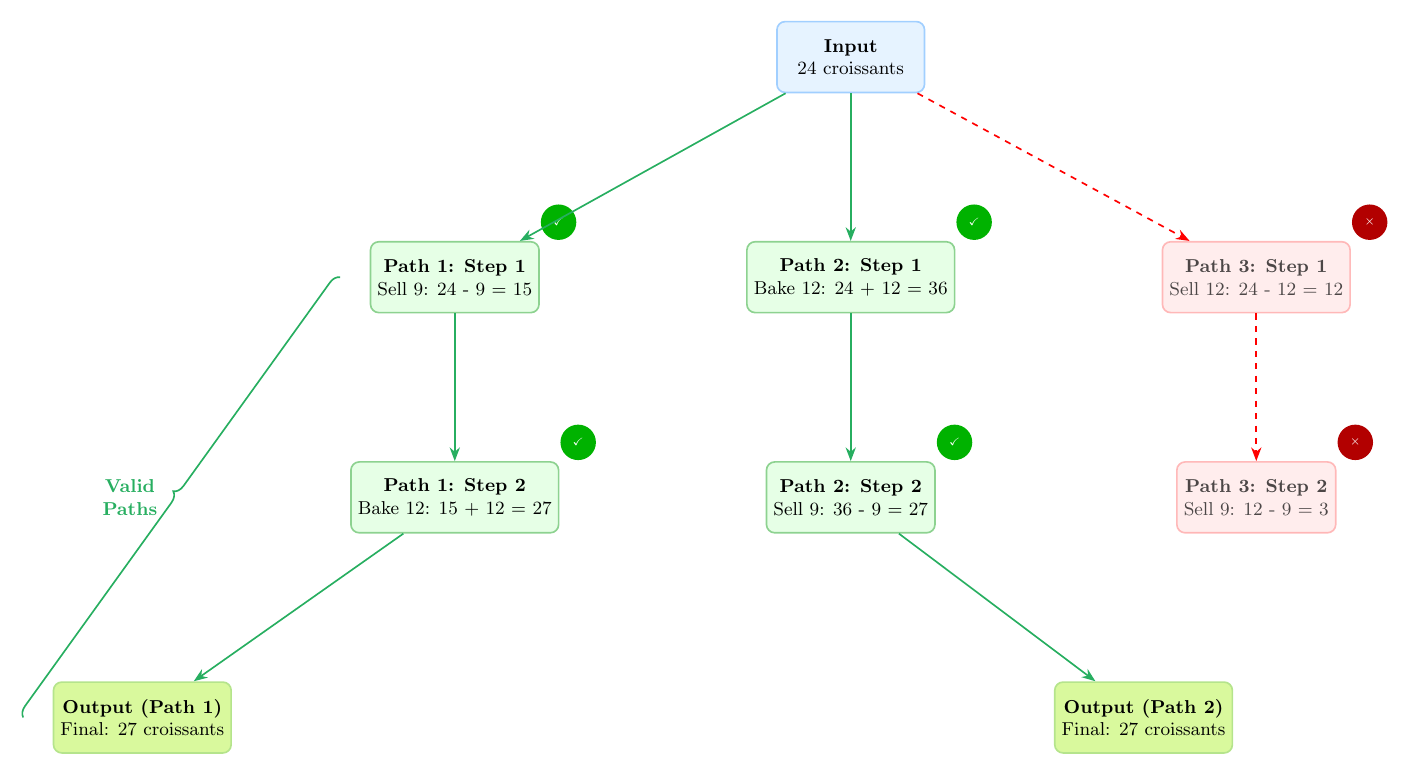}
   \caption{Tree-of-Thought (ToT): Branching and pruning}
        \label{fig:aot}
    \end{subfigure}
   \caption{Exploratory reasoning methodologies: ToT explores multiple paths with selective pruning, while AoT employs systematic backtracking to explore alternative approaches.}
    \label{fig:aot_tot}
\end{figure}


\newpage

\section{Overall Accuracy Score}\label{appendix: accuracy}

\begin{figure}[htbp]
  \centering
  \begin{subfigure}{0.8\columnwidth} 
    \centering    \includegraphics[width=\linewidth]{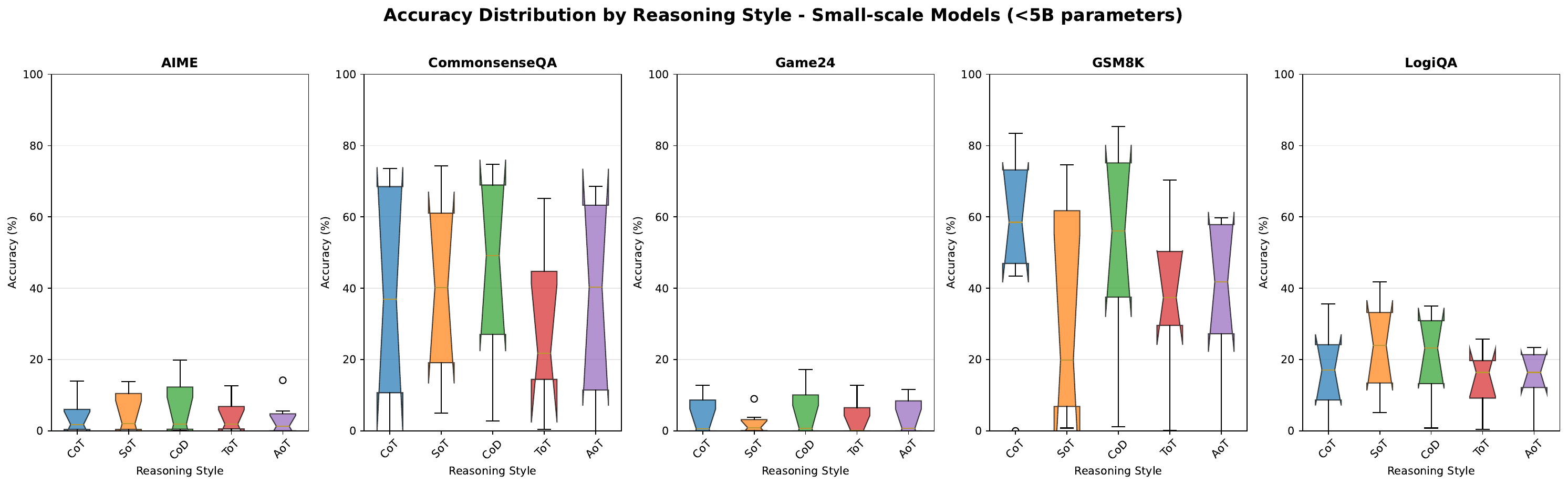}
    \caption{Accuracy boxplot for small models}
    \label{fig: small model boxplot}
  \end{subfigure}
  
  
  \begin{subfigure}{0.8\columnwidth} 
    \centering
    \includegraphics[width=\linewidth]{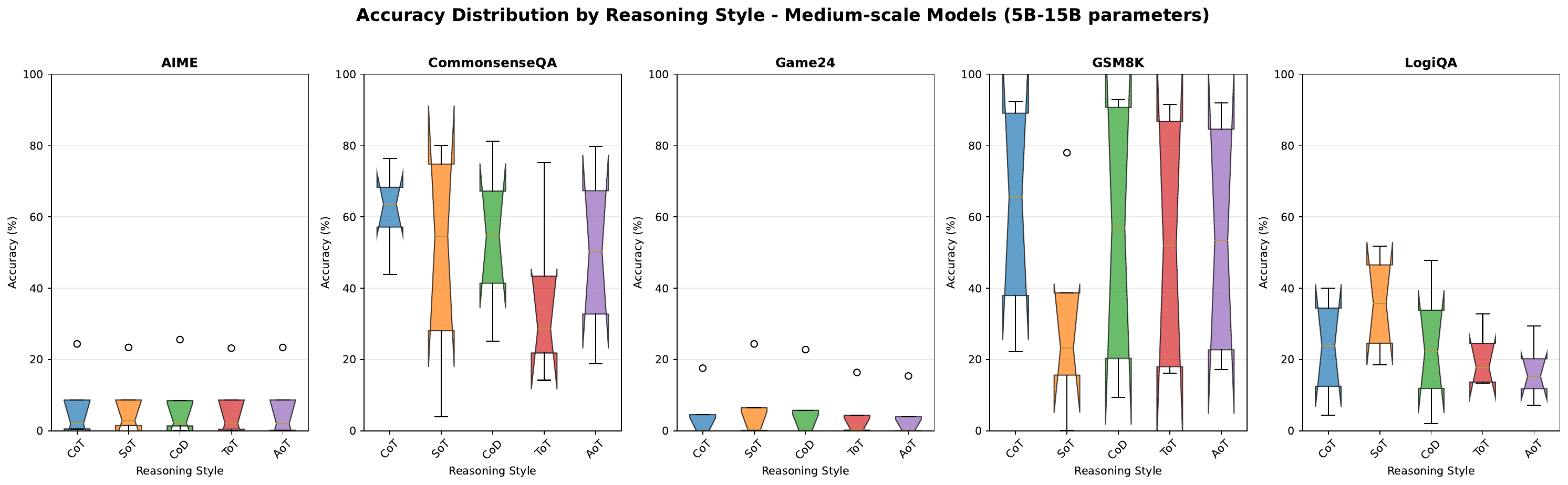}
    \caption{Accuracy Heatmap for medium models}
    \label{fig: medium model boxplot}
  \end{subfigure}


  \begin{subfigure}{0.8\columnwidth} 
    \centering
    \includegraphics[width=\linewidth]{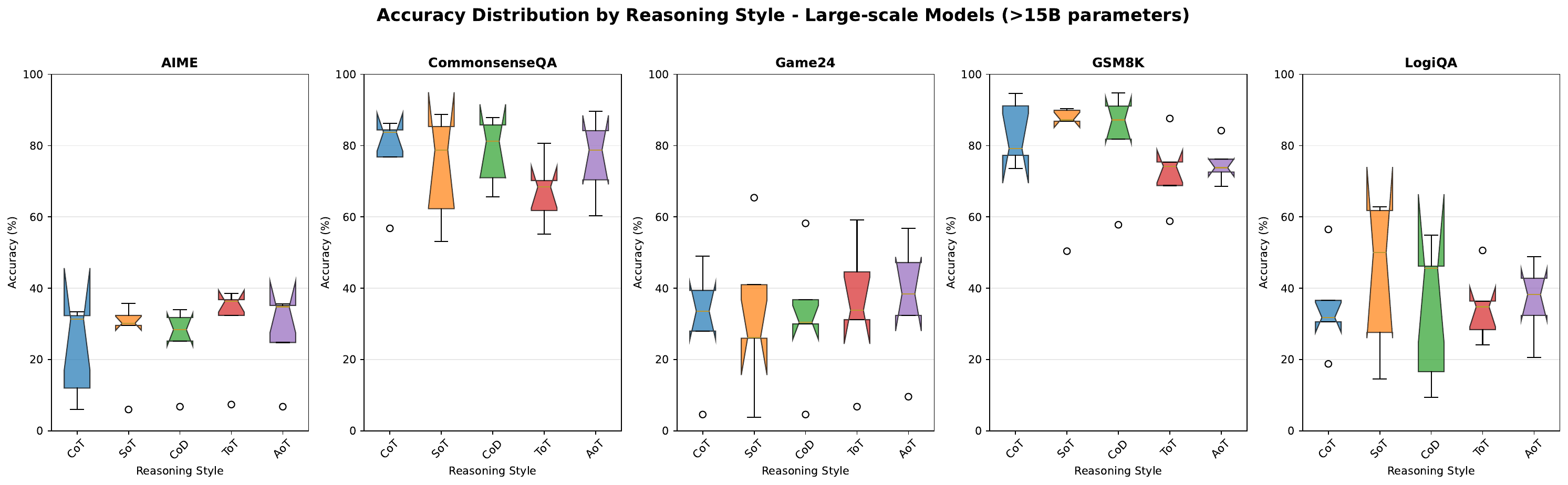}
    \caption{Accuracy Heatmap for small models}
    \label{fig: large model boxplot}
  \end{subfigure}

  \caption{Boxplot for three groups of model accuracy.}
  \label{fig: overall boxplot}
\end{figure}

\begin{figure}[htbp]  
  \centering
  
  \begin{subfigure}{\columnwidth}  
    \centering
    \includegraphics[width=\linewidth, height=0.28\textheight, keepaspectratio]{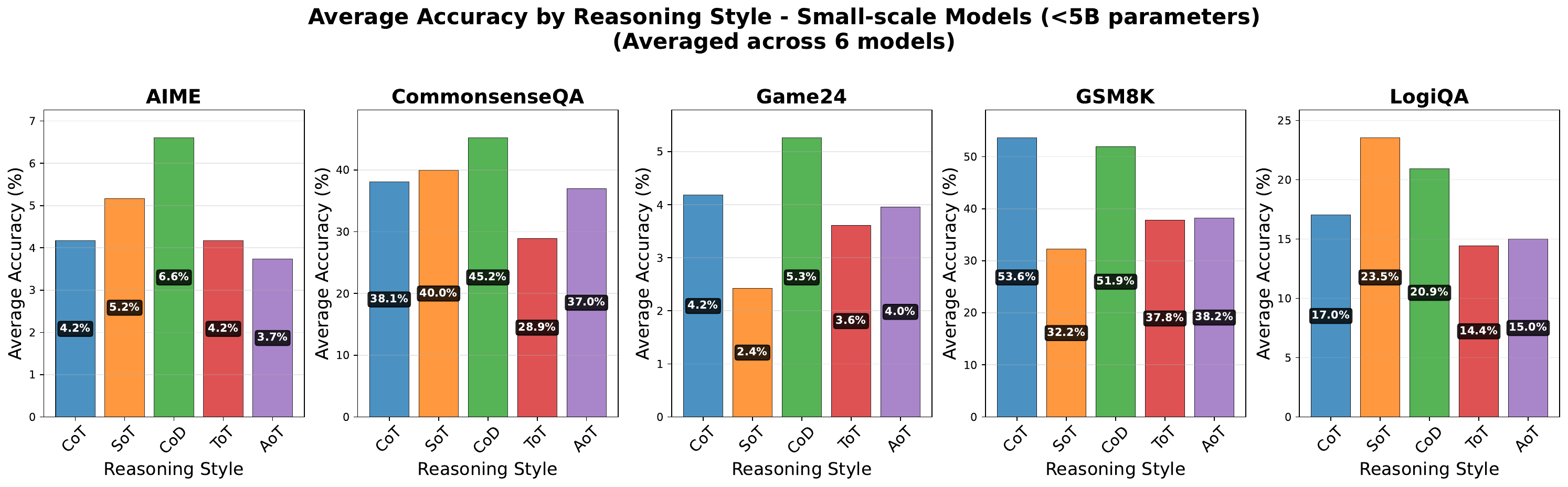}
    \caption{Small models ($<5$B parameters): averaged across 6 models including Gemma-270M, Qwen-0.5B, DeepSeek-R1-Distill-Qwen-1.5B, Gemma-2B, Qwen-3B, and Phi-3-Mini-3.8B.}
    \label{fig:small_model_barplot}
  \end{subfigure}
  \newpage
  \vspace{10pt}  
  
  \begin{subfigure}{\columnwidth}
    \centering
    \includegraphics[width=\linewidth, height=0.28\textheight, keepaspectratio]{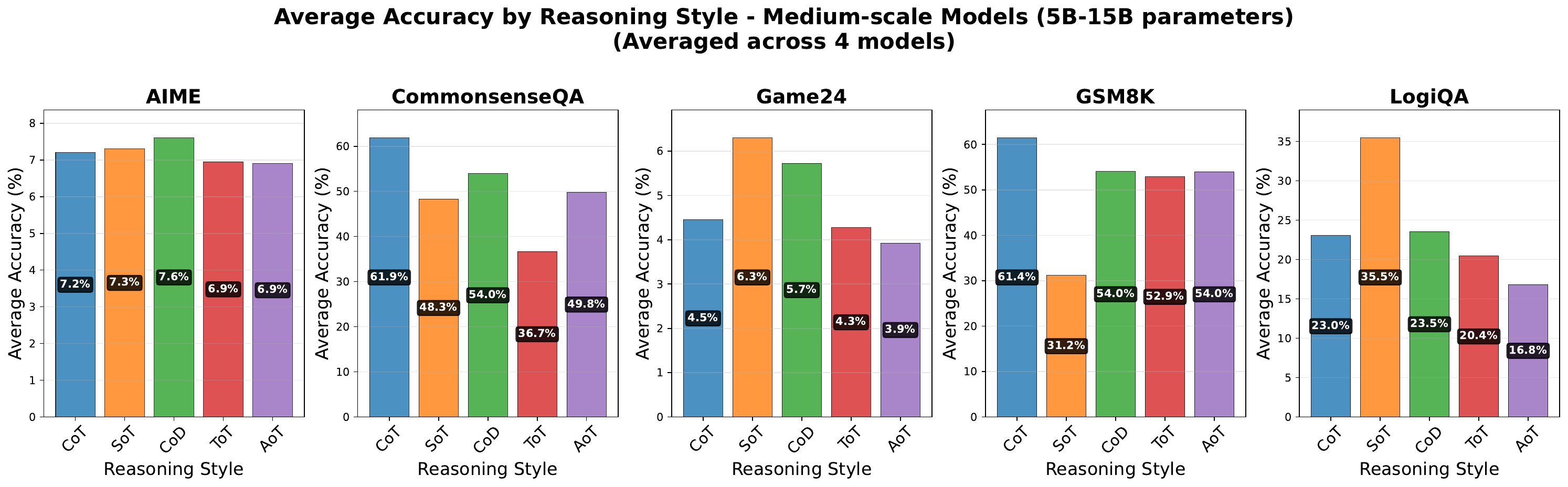}
    \caption{Medium models (5B-15B parameters): averaged across 4 models including Mistral-7B, Qwen-7B, Llama-3-8B, and Gemma-2-9B.}
    \label{fig:medium_model_barplot}
  \end{subfigure}
  
  \vspace{10pt}
  
  \begin{subfigure}{\columnwidth}
    \centering
    \includegraphics[width=\linewidth, height=0.28\textheight, keepaspectratio]{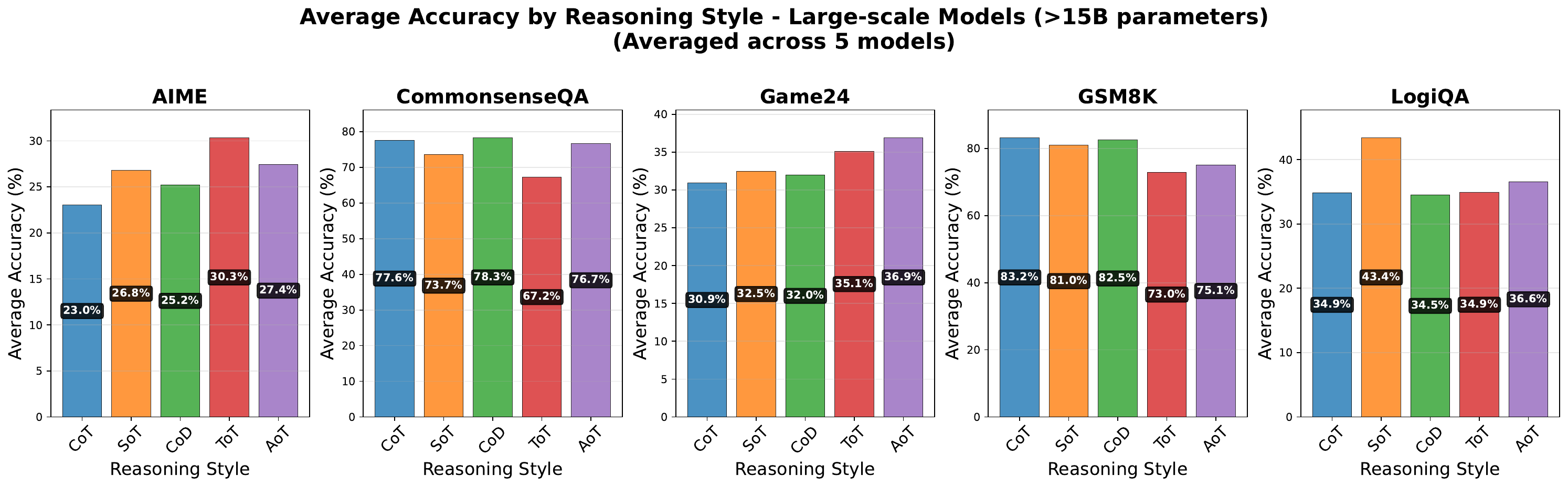}
    \caption{Large models ($>$15B parameters): averaged across 5 models including GPT-OSS-20B, Qwen-32B, Llama-3-70B, Qwen-72B, and GPT-OSS-120B.}
    \label{fig:large_model_barplot}
  \end{subfigure}
  
  \caption{Scaling behavior of reasoning styles across model sizes. Performance averaged within each size category across five reasoning tasks (AIME, CommonsenseQA, Game24, GSM8K, LogiQA). Key findings: (1) Complex search-based styles (ToT, AoT) show performance gains only in large models and can degrade performance in smaller models. (2) Concise styles (CoD, SoT) maintain more stable performance across scales, particularly excelling on knowledge retrieval and symbolic reasoning tasks. (3) Chain-of-Thought (CoT) demonstrates the most consistent baseline performance across all model scales for structured mathematical reasoning.}
  \label{fig:overall_accuracy_barplot}
\end{figure}

\begin{figure}[htbp]
    \centering
    \includegraphics[width=0.8\linewidth]{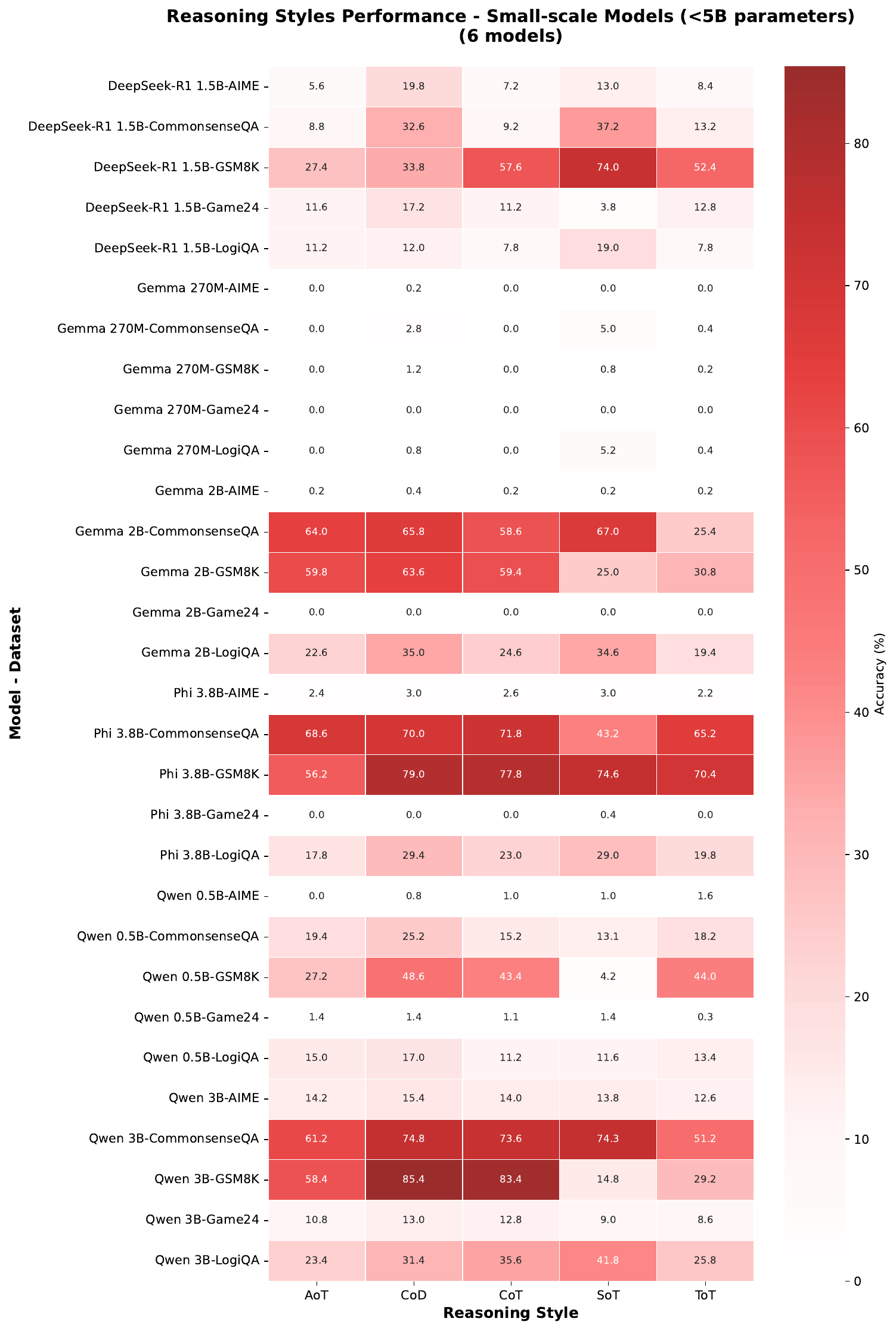}
    \caption{Accuracy Heatmap for small models}
    \label{fig: small model heatmap}
\end{figure}

\begin{figure}[htbp]
    \centering
    \includegraphics[width=\linewidth]{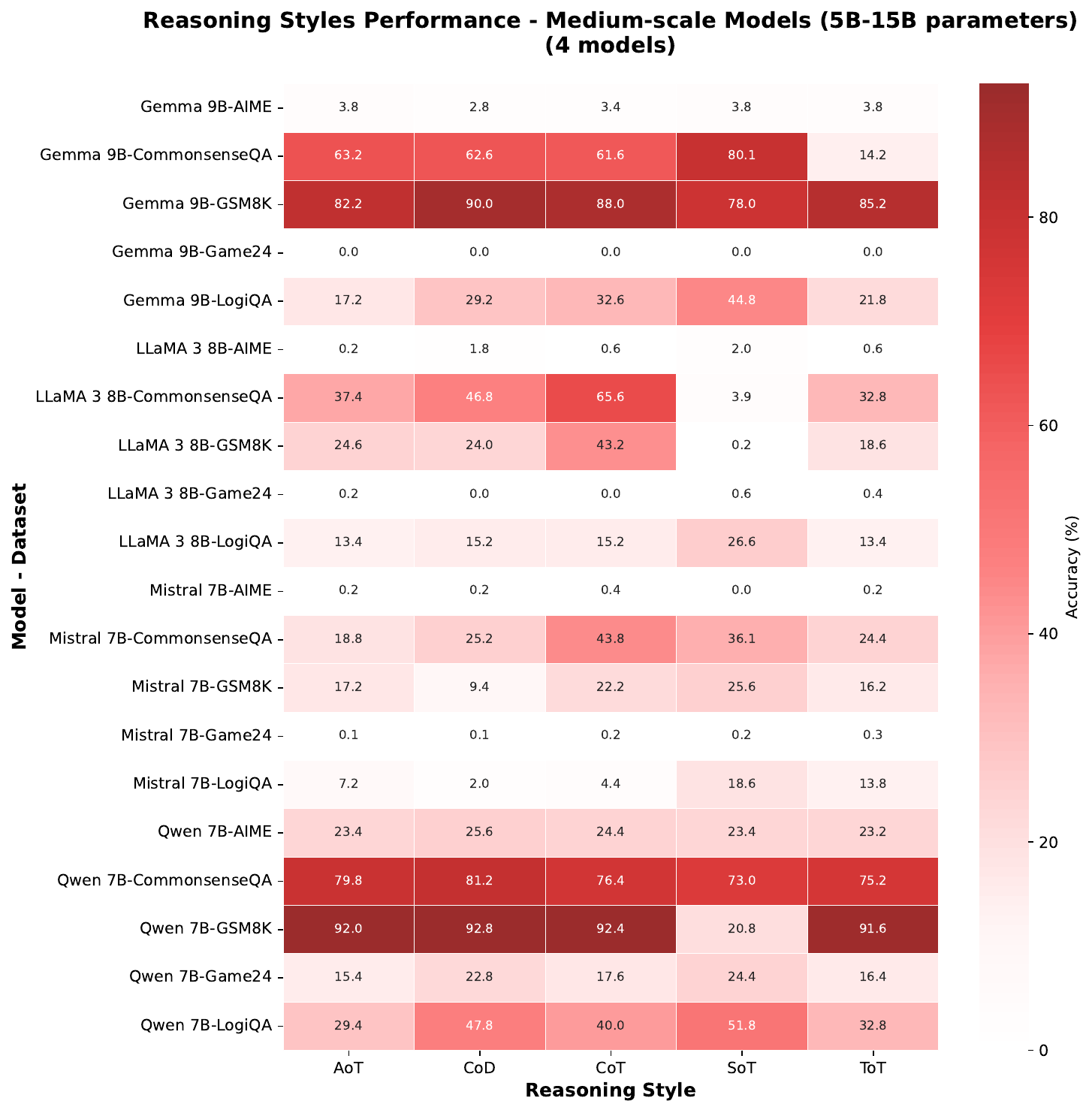}
    \caption{Accuracy Heatmap for medium models}
    \label{fig: medium model heatmap}
\end{figure}

\begin{figure}[htbp]
    \centering
    \includegraphics[width=0.9\linewidth]{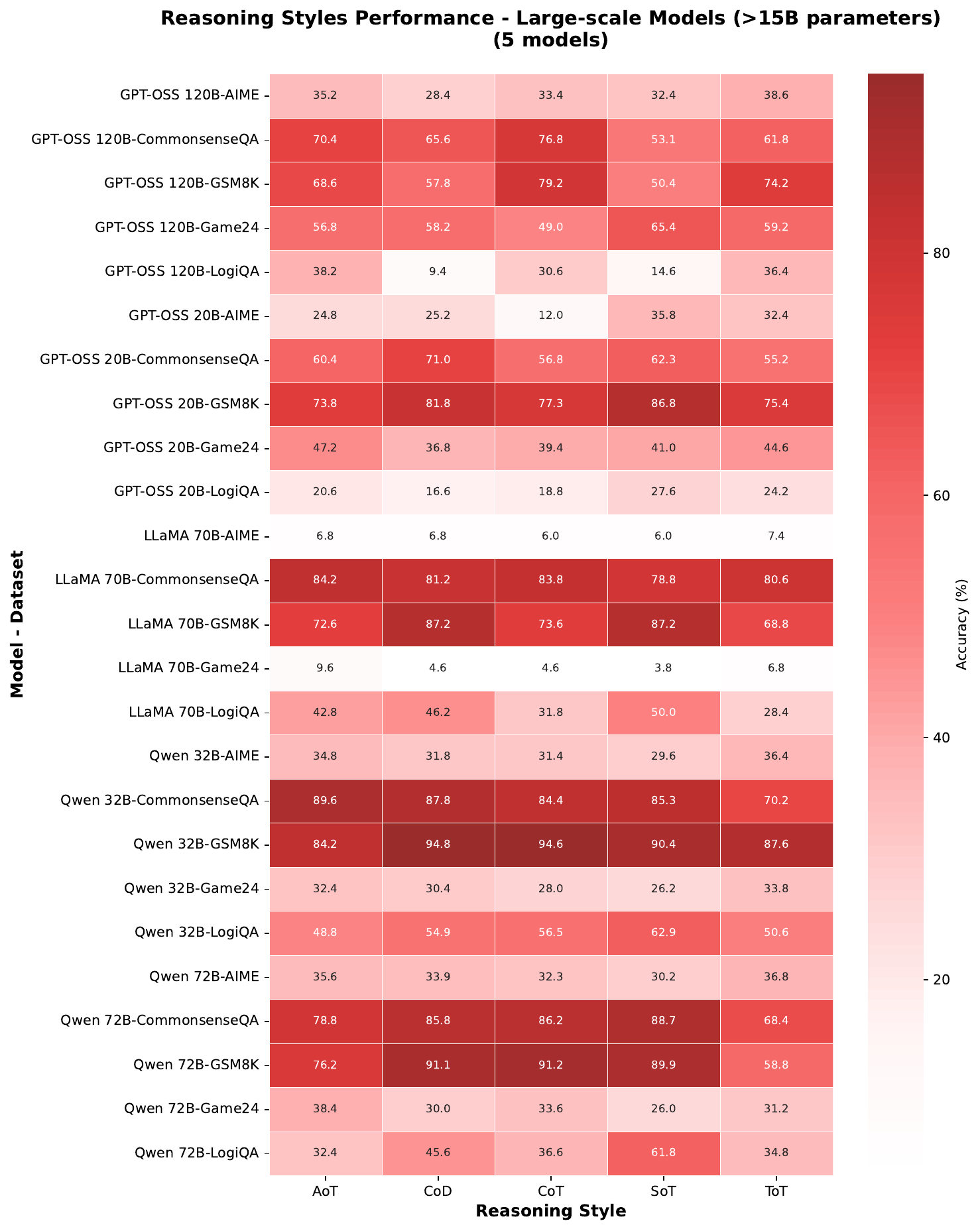}
    \caption{Accuracy Heatmap for large models}
    \label{fig: large model heatmap}
\end{figure}

\newpage

\section{Sample Prompts by Reasoning Style}\label{appendix: sys prompt}
Here we provide the sample prompts for each thinking style and we choose the CommonsenseQA dataset for example. 

\begin{tcolorbox}[colback=gray!10, colframe=black, title=Sample CommonsenseQA Problem]
Bob the lizard lives in a warm place with lots of water. Where does he probably live?
Options:
A) rock
B) tropical rainforest
C) jazz club
D) new mexico
E) rocky places
\end{tcolorbox}

\begin{tcolorbox}[colback=blue!10, colframe=blue!50, title=Chain of Thought (CoT), breakable]
\textbf{System:}
You are a careful math solver. Think step by step and show all your work clearly.
Explain your reasoning process, then provide your final answer.
At the end, put your final numerical answer in $\backslash$boxed$\{<\text{answer}>\}$ format.

\textbf{User:}
Problem:
Bob the lizard lives in a warm place with lots of water. Where does he probably live?
Options:
A) rock
B) tropical rainforest
C) jazz club
D) new mexico
E) rocky places

Let's work through this step by step, showing all calculations.

Please solve this step by step, showing all your work and reasoning.
Put your final numerical answer in $\backslash$boxed$\{<\text{think}>\}$ at the very end.
\end{tcolorbox}

\begin{tcolorbox}[colback=green!10, colframe=green!50, title=Tree of Thoughts (ToT), breakable]
\textbf{System:}
Imagine three different experts are solving this math problem together.
Each expert contributes their reasoning step by step. They discuss, check each other's work, and collaborate.
Show this collaborative problem-solving process with clear reasoning.
At the end, put the final agreed-upon answer in $\backslash$boxed$\{<\text{think}>\}$ format.

\textbf{User:}
Problem:
Bob the lizard lives in a warm place with lots of water. Where does he probably live?
Options:
A) rock
B) tropical rainforest
C) jazz club
D) new mexico
E) rocky places

Show how the three experts would collaborate on this problem, with each contributing their reasoning and checking each other's work.

Please solve this step by step, showing all your work and reasoning.
Put your final numerical answer in $\backslash$boxed$\{<\text{think}>\}$ at the very end.
\end{tcolorbox}

\begin{tcolorbox}[colback=orange!10, colframe=orange!50, title=Algorithm of Thoughts (AoT), breakable]
\textbf{System:}
Use two-phase reasoning to solve this problem:
Phase 1 - Forward Analysis: Think step by step through the problem from start to finish.
Phase 2 - Verification: Work backward from your answer to double-check it's correct.
Show both phases of your reasoning clearly.
At the end, put your final answer in $\backslash$boxed$\{<\text{think}>\}$ format.

\textbf{User:}
Problem:
Bob the lizard lives in a warm place with lots of water. Where does he probably live?
Options:
A) rock
B) tropical rainforest
C) jazz club
D) new mexico
E) rocky places

Use both forward analysis and backward verification. Show both phases of your reasoning clearly.

Please solve this step by step, showing all your work and reasoning.
Put your final numerical answer in $\backslash$boxed$\{<\text{think}>\}$ at the very end.
\end{tcolorbox}

\begin{tcolorbox}[colback=purple!10, colframe=purple!50, title=Chain of Draft (CoD), breakable]
\textbf{System:}
Use a Draft$\to$Refine workflow to solve problems step by step. Show your draft solution, then refine and improve it with detailed reasoning. 
At the end, put your final answer in $\backslash$boxed$\{<\text{think}>\}$ format.

\textbf{User:}
2015 is coming in 36 hours. What is the date one week from today in MM/DD/YYYY?

\textbf{Assistant:}$\{<\text{think}>\}$ 36h = 1.5 days; today: 2015/01/01 - 1.5 = 12/30/2014; 12/30/2014 + 7 = 01/06/2015. $\{</\text{think}>\}$
$\backslash$boxed\{01/06/2015\}

...

[22 more similar examples with $\{<\text{think}>\}$ format responses]

...

\textbf{User:} Bob the lizard lives in a warm place with lots of water. Where does he probably live?
Options:
A) rock
B) tropical rainforest
C) jazz club
D) new mexico
E) rocky places

Please solve this step by step, showing all your work and reasoning.
Put your final numerical answer in $\backslash$boxed$\{<\text{think}>\}$ at the very end.
\end{tcolorbox}

\begin{tcolorbox}[colback=red!10, colframe=red!50, title=Symbolic of Thoughts (SoT), breakable]
\textbf{System:}   

\textbf{Role \& Objective}   

You are a reasoning expert specializing in \textbf{Chunked Symbolism}, a cognitive reasoning technique that organizes numerical reasoning into structured steps. Your goal is to \textbf{utilize chunked symbolism} by representing information through \textbf{equations, variables, and step-by-step arithmetic}, while using minimal words.

Chunked Symbolism is inspired by the cognitive science principle of \textbf{chunking}—the idea that humans process information more efficiently when grouped into meaningful units. Instead of solving problems in a free-form manner, \textbf{Chunked Symbolism breaks down complex operations into smaller, structured steps}.

This method is particularly effective for:

- \textbf{Mathematical problems} (arithmetic, algebra, physics, engineering)

- \textbf{Symbolic reasoning} (logic-based computations, formula derivations)

- \textbf{Technical calculations} (financial modeling, physics simulations, unit conversions)

\textbf{How to Apply Chunked Symbolism}   

\textbf{Step-by-Step Guide}   

1. \textbf{Identify Variables} – Extract relevant numerical values and define variables.    

2. \textbf{Write Equations} – Represent the solution using \textbf{explicit mathematical formulas}.

3. \textbf{Perform Step-by-Step Computations} – Solve in \textbf{small, logical steps}, keeping each line clear.

4. \textbf{Label Units} – Maintain \textbf{consistent unit representation} to prevent ambiguity.   

5. \textbf{Final Answer Formatting} – Present the answer in the \textbf{provided format} for clarity.

\textbf{Rules \& Directives}   

1. \textbf{Use Equations \& Variables}    

   - Define variables before computation.   
   
   - Always use \textbf{explicit equations} to represent reasoning.

2. \textbf{Avoid Redundant Text}   

   - \textbf{Do not} restate the problem; go directly to calculations.   
   
   - Use \textbf{minimal context} only if it aids understanding.

3. \textbf{Apply Step-by-Step Arithmetic}   

   - Break operations into \textbf{small, structured steps}.   
   
   - Ensure each line contains only \textbf{one computation} for clarity.

4. \textbf{Output Format}   

   - Use the exact structured format:   
   
   ```   
   
 $<\text{think}>$
   [shorthand reasoning]
$</\text{think}>$
   $\backslash$boxed\{[Final answer]\}
   ```
   - The \textbf{final answer must be boxed}.
   - \textbf{If the question is multiple-choice, return the correct letter option inside the box.}
   - \textbf{Use minimal words in your response.}

\textbf{User:}
A car accelerates at 2.5 m/s$^2$ for 10 seconds. If its initial velocity was 15 m/s, what is its final velocity?

\textbf{Assistant:} $<\text{think}>$
a = 2.5 m/s$^2$
t = 10 s
vi = 15 m/s
vf = 15 + (2.5 × 10)
vf = 40 m/s
$</\text{think}>$
$\backslash$boxed\{40\}

\textbf{User:} If a product costs \$120 and there is a 15\% discount, what is the final price?
Choices:
A) \$10
B) \$97
C) 102

\textbf{Assistant:}$<\text{think}>$
op = 120
d = 15\%
dp = 120 × (15 / 100) = 18
fp = 120 - 18 = 102
$</\text{think}>$
$\backslash$boxed\{C\}

\textbf{User:} Question: A circuit has a voltage of 12V and a resistance of 4$\Omega$. What is the current?

\textbf{Assistant:} $<\text{think}>$
V = 12V
R = 4$\Omega$
I = 12 / 4 = 3A
$</\text{think}>$
$\backslash$boxed\{3\}

\textbf{User:} Bob the lizard lives in a warm place with lots of water. Where does he probably live?
Options:
A) rock
B) tropical rainforest
C) jazz club
D) new mexico
E) rocky places

Please solve this step by step, showing all your work and reasoning.
Put your final numerical answer in $\backslash$boxed$\{<\text{think}>\}$ at the very end.
\end{tcolorbox}

\section{Cross-Model Comparison}
\subsection{Case Study 1: Game24}\label{case study 1}

\begin{table}[ht]
\centering
\small
\caption{Final responses from different models and reasoning methods for Game24 problem "2, 2, 7, 10"}
\label{tab:game24_responses}
\begin{tabular}{llllll}
\toprule
\textbf{Model} & \textbf{CoT} & \textbf{ToT} & \textbf{AoT} & \textbf{CoD} & \textbf{SoT} \\
\midrule
\textbf{Qwen 3B} & No solution $\checkmark$ & No solution $\checkmark$ & No solution $\checkmark$ & No answer $\times$ & No solution $\checkmark$ \\
\textbf{Qwen 7B} & No answer $\times$ & No answer $\times$ & No answer $\times$ & No answer $\times$ & No answer $\times$ \\
\textbf{Qwen 32B} & Invalid expr. $\times$ & No answer $\times$ & No answer $\times$ & No answer $\times$ & No answer $\times$ \\
\textbf{Qwen 72B} & \textbf{Valid sol.} $\checkmark$ & No solution $\checkmark$ & \textbf{Valid sol.} $\checkmark$ & Invalid expr. $\times$ & No answer $\times$ \\
\textbf{GPT-OSS 120B} & No answer $\times$ & No answer $\times$ & No answer $\times$ & No answer $\times$ & \textbf{Valid sol.} $\checkmark$ \\
\bottomrule
\end{tabular}
\end{table}

\begin{table}[ht]
\centering
\small
\caption{Success rate analysis across models and methods}
\label{tab:success_rates}
\begin{tabular}{lrrrrr}
\toprule
\textbf{Model} & \textbf{Valid Solutions} & \textbf{Correct "No Solution"} & \textbf{Invalid Solutions} & \textbf{No Answer} & \textbf{Success Rate} \\
\midrule
Qwen 3B & 0 & 4 & 0 & 1 & 80\% \\
Qwen 7B & 0 & 0 & 0 & 5 & 0\% \\
Qwen 32B & 0 & 0 & 1 & 4 & 0\% \\
Qwen 72B & 2 & 1 & 1 & 1 & 60\% \\
GPT-OSS 120B & 1 & 0 & 0 & 4 & 20\% \\
\midrule
\textbf{Total} & \textbf{3} & \textbf{5} & \textbf{2} & \textbf{15} & \textbf{32\%} \\
\bottomrule
\end{tabular}
\end{table}

\begin{table}[ht]
\centering
\small
\caption{Method performance across all models}
\label{tab:method_analysis}
\begin{tabular}{lrrrrr}
\toprule
\textbf{Method} & \textbf{Valid Solutions} & \textbf{Correct "No Solution"} & \textbf{Invalid Solutions} & \textbf{No Answer} & \textbf{Success Rate} \\
\midrule
CoT & 2 & 2 & 1 & 0 & 80\% \\
ToT & 0 & 2 & 0 & 3 & 40\% \\
AoT & 2 & 2 & 0 & 1 & 80\% \\
CoD & 0 & 0 & 1 & 4 & 0\% \\
SoT & 1 & 1 & 0 & 3 & 40\% \\
\bottomrule
\end{tabular}
\end{table}

\begin{table}[ht]
\centering
\small
\caption{Valid solutions discovered}
\label{tab:valid_solutions}
\begin{tabular}{llp{6cm}}
\toprule
\textbf{Model} & \textbf{Method} & \textbf{Expression and Verification} \\
\midrule
Qwen 72B & CoT & $(10 \div 2 + 7) \times 2 = (5 + 7) \times 2 = 12 \times 2 = 24$ \\
Qwen 72B & AoT & $(10 \div 2 + 7) \times 2 = (5 + 7) \times 2 = 12 \times 2 = 24$ \\
GPT-OSS 120B & SoT & $(7 + 10/2) \times 2 = (7 + 5) \times 2 = 12 \times 2 = 24$ \\
\bottomrule
\end{tabular}
\end{table}
\begin{table}[ht]
\centering
\small
\caption{Detailed final expressions from each model-method combination}
\label{tab:game24_expressions}
\begin{tabular}{p{2.5cm}p{2.5cm}p{8cm}}
\toprule
\textbf{Model} & \textbf{Method} & \textbf{Final Expression} \\
\midrule
Qwen 3B 
& CoT & $\boxed{\text{No solution}}$ \\
& ToT & $\boxed{\text{No solution}}$ \\
& AoT & $\boxed{\text{No solution}}$ \\
& CoD & No final boxed answer (infinite loop) \\
& SoT & $\boxed{\text{No solution}}$ \\
\midrule
Qwen 7B 
& CoT & No final boxed answer (infinite loop) \\
& ToT & No final boxed answer (infinite loop) \\
& AoT & No final boxed answer (infinite loop) \\
& CoD & No final boxed answer (infinite loop) \\
& SoT & No final boxed answer (infinite loop) \\
\midrule
Qwen 32B 
& CoT & $\boxed{(10 + 2) \times 2 + 7 - 7}$ (uses 7 twice) \\
& ToT & No final boxed answer (infinite loop) \\
& AoT & No final boxed answer (infinite loop) \\
& CoD & No final boxed answer (infinite loop) \\
& SoT & No final boxed answer (infinite loop) \\
\midrule
Qwen 72B 
& CoT & $\boxed{(10 \div 2 + 7) \times 2}$ \textbf{(Valid solution)} \\
& ToT & $\boxed{\text{No solution}}$ \\
& AoT & $\boxed{(10 \div 2 + 7) \times 2}$ \textbf{(Valid solution)} \\
& CoD & $\boxed{(2 + 2) \times (10 - 7) + 10 + 2}$ (number reuse) \\
& SoT & No final boxed answer (infinite loop) \\
\midrule
GPT-OSS 120B 
& CoT & No final boxed answer (stream of consciousness) \\
& ToT & No final boxed answer (chaotic exploration) \\
& AoT & No final boxed answer (endless calculations) \\
& CoD & No final boxed answer (partial insight) \\
& SoT & $\boxed{(7 + 10/2) \times 2}$ \textbf{(Valid solution)} \\
\bottomrule
\end{tabular}
\end{table}

\newpage

\subsection{Case Study 2: AIME}\label{case study 2}
Let $n=2^{31}3^{19}.$ How many positive integer divisors of $n^2$ are less than $n$ but do not divide $n$?

\textbf{Ground Truth:} 589
\textbf{Legend:} 
\textcolor{green}{Green} = Correct (589), 
\textcolor{orange}{Orange} = Near-correct (588), 
\textcolor{red}{Red} = Incorrect
\begin{longtable}{|l|c|c|c|c|c|}
\hline
\textbf{Model} & \textbf{CoT} & \textbf{ToT} & \textbf{AoT} & \textbf{CoD} & \textbf{SoT} \\
\hline
\endhead

\textbf{Qwen3B} & \textcolor{red}{1816} & \textcolor{green}{589} & \textcolor{orange}{588} & \textcolor{red}{346} & \textcolor{orange}{588} \\
\hline

\textbf{Qwen7B} & \textcolor{green}{589} & \textcolor{orange}{588} & \textcolor{orange}{588} & \textcolor{green}{589} & \textcolor{red}{909} \\
\hline

\textbf{Qwen32B} & \textcolor{orange}{588} & \textcolor{orange}{588} & \textcolor{orange}{588} & \textcolor{orange}{588} & \textcolor{orange}{588} \\
\hline

\textbf{Qwen72B} & \textcolor{green}{589} & \textcolor{green}{589} & \textcolor{red}{908} & \textcolor{orange}{588} & \textcolor{green}{589} \\
\hline

\textbf{GPT-OSS-120B} & \textcolor{green}{589} & \textcolor{green}{589} & \textcolor{green}{589} & \textcolor{green}{589} & \textcolor{red}{908} \\
\hline

\textbf{Ground Truth} & \multicolumn{5}{|c|}{\textcolor{green}{589}} \\
\hline

\end{longtable}



\section{Experiment Settings}
\label{appendix:experiment_settings}

To ensure fair and reproducible comparisons across all models, we maintained consistent prompting configurations. We set the temperature to 0 for deterministic outputs and standardized both the maximum new tokens and model context length for each dataset. 

\begin{table}[h!]
\centering
\caption{Prompting Configuration for All Models}
\label{tab:experiment_settings}
\begin{tabular}{|l|c|c|c|}
\hline
\textbf{Dataset} & \textbf{Max New Tokens} & \textbf{Context Length} & \textbf{Task Characteristics} \\
\hline
AIME & 2048 & 8192 & Complex mathematical reasoning \\
CommonsenseQA & 512 & 8192 & Multiple-choice questions \\
Game24 & 2048 & 8192 & Mathematical puzzle solving \\
GSM8K & 512 & 8192 & Grade school math problems \\
LogiQA & 512 & 8192 & Logical reasoning tasks \\
\hline
\end{tabular}
\end{table}

\end{document}